\documentclass[letterpaper]{article} 
\usepackage[preprint]{aaai2027}  
\usepackage[hyphens]{url}  
\usepackage{graphicx} 
\usepackage{natbib}  
\usepackage{caption} 
\usepackage{amsmath}
\usepackage{amssymb}
\usepackage{multirow}

\usepackage{booktabs}
\usepackage{pifont}

\usepackage{algorithm}
\usepackage{algorithmic}
\usepackage{listings}
\usepackage[most]{tcolorbox}

\newsavebox{\promptbreakarrow}
\savebox{\promptbreakarrow}{\raisebox{0ex}[0ex][0ex]{\textcolor{black!45}{$\hookrightarrow$}}\space}

\lstdefinestyle{promptstyle}{
  basicstyle=\scriptsize\ttfamily,
  breaklines=true,
  breakindent=0pt,
  postbreak=\usebox{\promptbreakarrow},
  columns=fullflexible,
  keepspaces=true,
  frame=none,
  aboveskip=0pt,
  belowskip=0pt,
  showstringspaces=false
}
\definecolor{pbBlue}{HTML}{EEF4FB}
\definecolor{pbBlueEdge}{HTML}{C3D6EC}
\definecolor{pbGreen}{HTML}{F0F7EC}
\definecolor{pbGreenEdge}{HTML}{CFE3C2}
\definecolor{pbPurple}{HTML}{F3EEF8}
\definecolor{pbPurpleEdge}{HTML}{DDCFEB}
\definecolor{pbAmber}{HTML}{FEFAF0}
\definecolor{pbAmberEdge}{HTML}{F2DFAE}

\newtcolorbox{promptbox}[2][]{%
  breakable, enhanced,
  width=\columnwidth,
  colback=pbBlue, colframe=pbBlueEdge, colbacktitle=pbBlueEdge,
  coltitle=black, fonttitle=\footnotesize\bfseries,
  boxrule=0.6pt, arc=0.8mm,
  top=7pt, bottom=4pt, left=4pt, right=3pt,
  attach boxed title to top left={yshift=-2.2mm, xshift=3.5mm},
  boxed title style={boxrule=0pt, colframe=white, arc=0.6mm},
  title=#2, #1
}

\title{StreamOPD: A Post-Training Recipe with Spatio-Temporal Cue Gating for Streaming Video Understanding}

\author{
    Keming Wu\textsuperscript{1},
    Baoyi Wang\textsuperscript{2},
    Kaichen Zhang\textsuperscript{3},
    Xiang An\textsuperscript{4},
    Zuhao Yang\textsuperscript{5},
    Sudong Wang\textsuperscript{6},\\
    Haowei Zhu\textsuperscript{1},
    Tingxuan Huang\textsuperscript{1},
    Hongcheng Gao\textsuperscript{1},
    Bin Wang\textsuperscript{1}\corresponding
}

\affiliations{
    \textsuperscript{1}Tsinghua University,
    \textsuperscript{2}Zhejiang University,
    \textsuperscript{3}The University of Hong Kong,\\
    \textsuperscript{4}LMMs-Lab,
    \textsuperscript{5}Nanyang Technological University,\\
    \textsuperscript{6}Hong Kong University of Science and Technology (Guangzhou)\\
    Project page: \url{https://unix-ai-lab.github.io/StreamOPD}
}

\begin{document}

\maketitle

\begin{abstract}
Streaming video understanding demands direct responses from the causally observed prefix of an unfolding video. Existing systems add inference-time memory, retrieval, and compression, yet a training-free sliding-window baseline already matches them. We therefore fix a memory-free recent-window protocol and ask how far post-training alone can go. Reinforcement learning with verifiable rewards fits this regime poorly, encouraging long ``think-then-answer'' generations, while on-policy distillation (OPD) supplies dense token-level teacher supervision on student trajectories but is stable only when both models train in thinking mode. These observations lead to \textsc{StreamOPD}, a recipe combining verifiable streaming-video data, thinking-mode OPD, and instruct-mode deployment. It raises StreamingBench from $77.9\%$ to $83.9\%$---within $0.3$ points of the 9B teacher---and improves OVO-Bench excluding its hallucination-detection subtask (HLD) by $9.1$ points under unchanged inference. As a teacher-privilege extension, \emph{Spatio-Temporal CueGate (ST-CueGate)} aggregates cue-versus-no-cue teacher likelihood ratios into a group-relative response score that reweights OPD. It reaches $71.9\%$ on OVO-Bench (excluding HLD) and $64.9\%$ on Video-MME, and is the only variant that stays above the base model on all four benchmarks. Replacing the teacher with a frozen copy of the student's initial policy---on-policy self-distillation---retains most of these gains and lifts HLD to $57.0\%$, above both the untrained student and the 9B teacher, so abstention loss is not intrinsic to the recipe. We provide a transparent and reproducible reference for open-source streaming-video research.

\end{abstract}

\section{Introduction}

Streaming video understanding asks a multimodal large language model (MLLM) to answer questions while a video is still unfolding~\citep{lin2026streamingbench,niu2025ovo,chen2024videollm}. Unlike offline video QA, the model cannot look ahead or repeatedly revisit the full clip: it must respond directly from the causally observed prefix available when the question arrives. This regime is central to live assistants, wearable cameras, and monitored environments, yet remains difficult for open MLLMs. StreamingBench~\citep{lin2026streamingbench} and OVO-Bench~\citep{niu2025ovo} expose this gap through real-time perception, backward tracing, and forward-looking queries.

Most streaming methods address this challenge with memory banks, KV-cache compression, retrieval, or specialized streaming modules. SimpleStream~\citep{shen2026simple}, however, shows that a strong, training-free recent-window baseline can already match or surpass substantially more complex systems, reaching $67.7\%$ on OVO-Bench and $80.6\%$ on StreamingBench with only four recent frames. This finding suggests that architectural complexity is not necessarily the primary bottleneck. We therefore fix the same simple inference protocol throughout---no memory bank, retrieval, compression, or online reasoning module---and isolate a cleaner question: can \emph{post-training} close the small-to-large model gap while leaving the streaming inference path unchanged? Concurrent work brings reasoning into the stream and learns when to answer~\citep{liu2026thinking}; we instead move reasoning to training time and preserve direct instruct-mode deployment.

Modern post-training introduces a different tension. Reinforcement learning with verifiable rewards (RLVR)~\citep{lambert2024tulu,guo2025deepseek}, GRPO~\citep{shao2024deepseekmath}, and multimodal reasoning recipes~\citep{zhang2025openmmreasoner,yang2025longvt,yang2026paravt} elicit stronger reasoning through extended generations, but those reasoning tokens must be produced before the final answer and are therefore poorly matched to direct-response streaming. Our teacher-free GRPO baseline makes the mismatch concrete: it maintains a strong reward-compatible score while drifting toward long rationales that fail under the deployed direct-answer format. Sparse task reward constrains the final answer but not the trajectory that precedes it.

\begin{figure}[t]
\centering
\includegraphics[width=0.9\columnwidth]{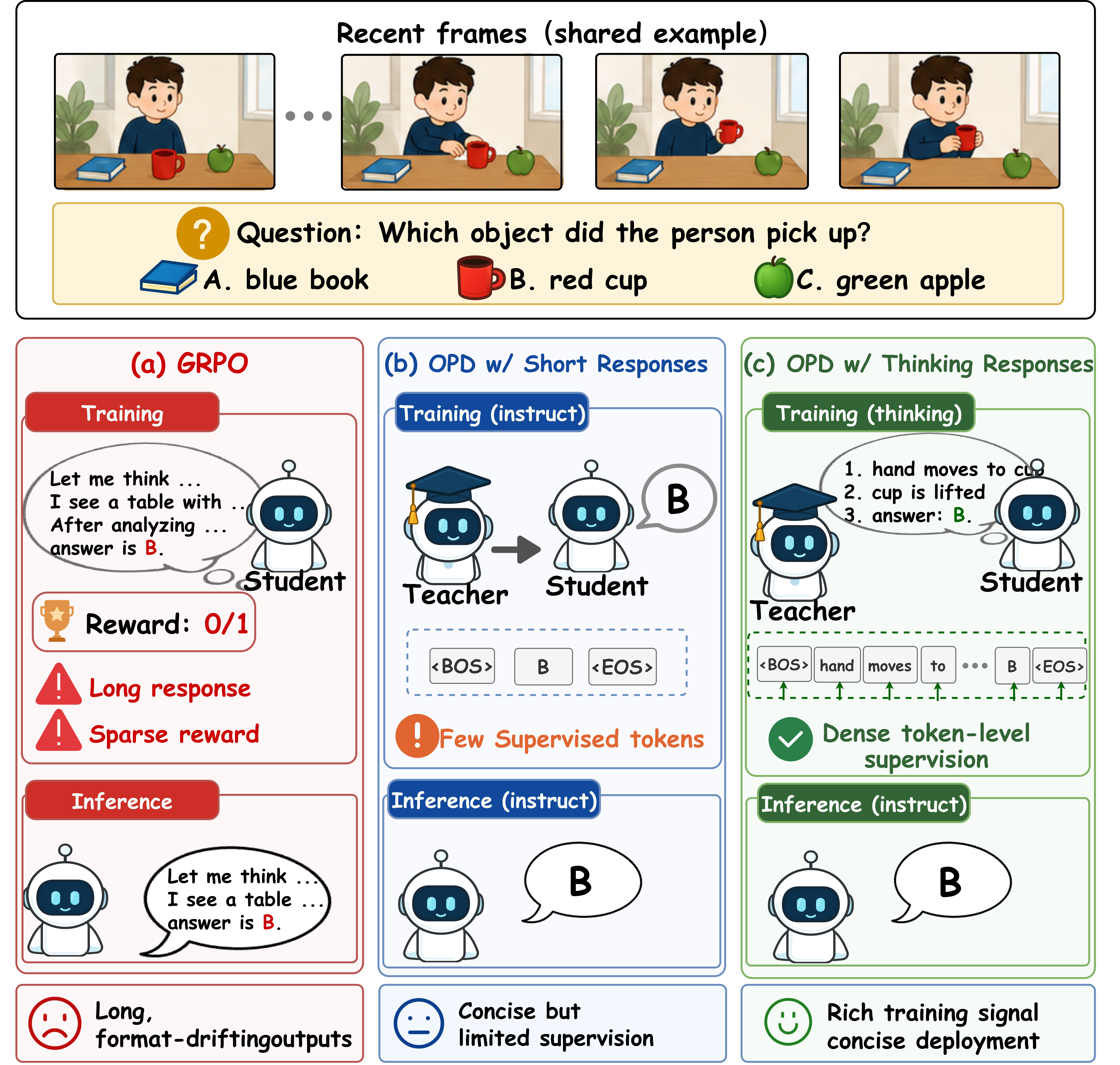}
\caption{Teacher/student training-mode pairings. All models are evaluated in instruct mode under the fixed recent-window protocol. Among the tested pairings, only both-thinking training converges in our runs and reaches $83.9\%$ on StreamingBench.}
\label{fig:quadrant}
\end{figure}

On-policy distillation (OPD) has recently emerged as an alternative that evaluates a teacher on student-generated trajectories and supplies dense token-level supervision~\citep{agarwal2024policy,gu2024minillm,zhao2026self}. In our mode comparison, both-instruct and teacher-thinking/student-instruct training collapse early, whereas both-thinking training converges. The failed runs combine one- or few-token responses with substantially larger gradient norms, which is consistent with a short-trajectory scale hypothesis. Response length is not independently controlled, however, so mode-specific policy distributions, teacher--student KL, estimator variance, normalization, and clipping remain competing explanations. We therefore treat length as a mechanistic hypothesis and the thinking-train/instruct-infer pairing as the empirical recipe supported by our experiments (Fig.~\ref{fig:quadrant}).

Under this recipe, standard OPD raises a Qwen3.5-4B~\citep{qwen35blog} student from $77.9\%$ to $83.9\%$ on StreamingBench---within $0.3$ points of a Qwen3.5-9B teacher---without changing inference, and gains $9.1$ points on OVO-Bench excluding its hallucination-detection (HLD) subtask, which scores abstention on unanswerable queries rather than streaming event recall. These gains motivate a second question: beyond choosing a teacher, what information should the teacher receive? We study teacher inputs as \emph{privileged information}, distinguishing signals derived from the same underlying training clip from ungrounded manipulations of the learning signal. Passive teacher-only visual cues do not consistently outperform standard OPD. Cue-versus-no-cue scoring shows a sparse teacher response: $56.5\%$ of token-level contrasts are near zero, while the top $20\%$ of positive-score tokens account for $82\%$ of the total positive mass. We therefore propose \emph{Spatio-Temporal CueGate (ST-CueGate)}, a nested-context gate that aggregates these contrasts into a group-relative, response-level likelihood proxy while leaving the student and inference path unchanged. ST-CueGate reaches $84.6\%$ on StreamingBench, $71.9\%$ on OVO-Bench (excluding HLD), $64.9\%$ on Video-MME, and $61.4\%$ on LongVideoBench.
We use \textsc{StreamOPD} for the complete recipe and ST-CueGate for its optional teacher-privilege extension.

Our main contributions are:

\begin{itemize}
\item We document two post-training mismatches in streaming video: teacher-free GRPO optimizes its task reward while drifting toward increasingly verbose responses, whereas the tested OPD configurations involving instruct-mode training collapse.
\item We establish StreamOPD, a practical recipe that combines a reproducible 25k verifiable-data pipeline, the empirically stable thinking-train/instruct-infer OPD configuration, and a fixed memory-free evaluation protocol. On top of this core, ST-CueGate uses cue-versus-no-cue teacher likelihood ratios to reweight student responses without modifying deployment.
\item We report cross-benchmark gains under unchanged inference: StreamOPD improves over the untrained student on StreamingBench and OVO-Bench, while ST-CueGate also improves both general-video benchmarks.
\end{itemize}

Together, these results indicate that a fixed streaming inference path still leaves substantial headroom for post-training alone to recover, and that what the teacher is shown can matter as much as how large the teacher is.

\section{Method}

\textsc{StreamOPD} consists of a core post-training recipe and an optional teacher-privilege extension. The core fixes the recent-window deployment protocol, constructs verifiable training data, trains student and teacher in thinking mode, and deploys the student in instruct mode. This design follows two empirical failures: teacher-free GRPO drifts toward long, deployment-incompatible trajectories, whereas the tested OPD configurations involving instruct-mode training collapse.

On top of this core recipe, we ask not only which teacher to use, but how the teacher responds to privileged information on each student-generated trajectory. Our cue-versus-no-cue analysis shows that the teacher's likelihood shift is sparse and highly concentrated, so uniform cue conditioning does not distinguish trajectories with different cue sensitivity. We therefore introduce ST-CueGate as an extension that aggregates token-level conditional likelihood ratios into a group-relative response proxy for reweighting OPD.

\begin{figure*}[t]
\centering
\includegraphics[width=0.95\textwidth]{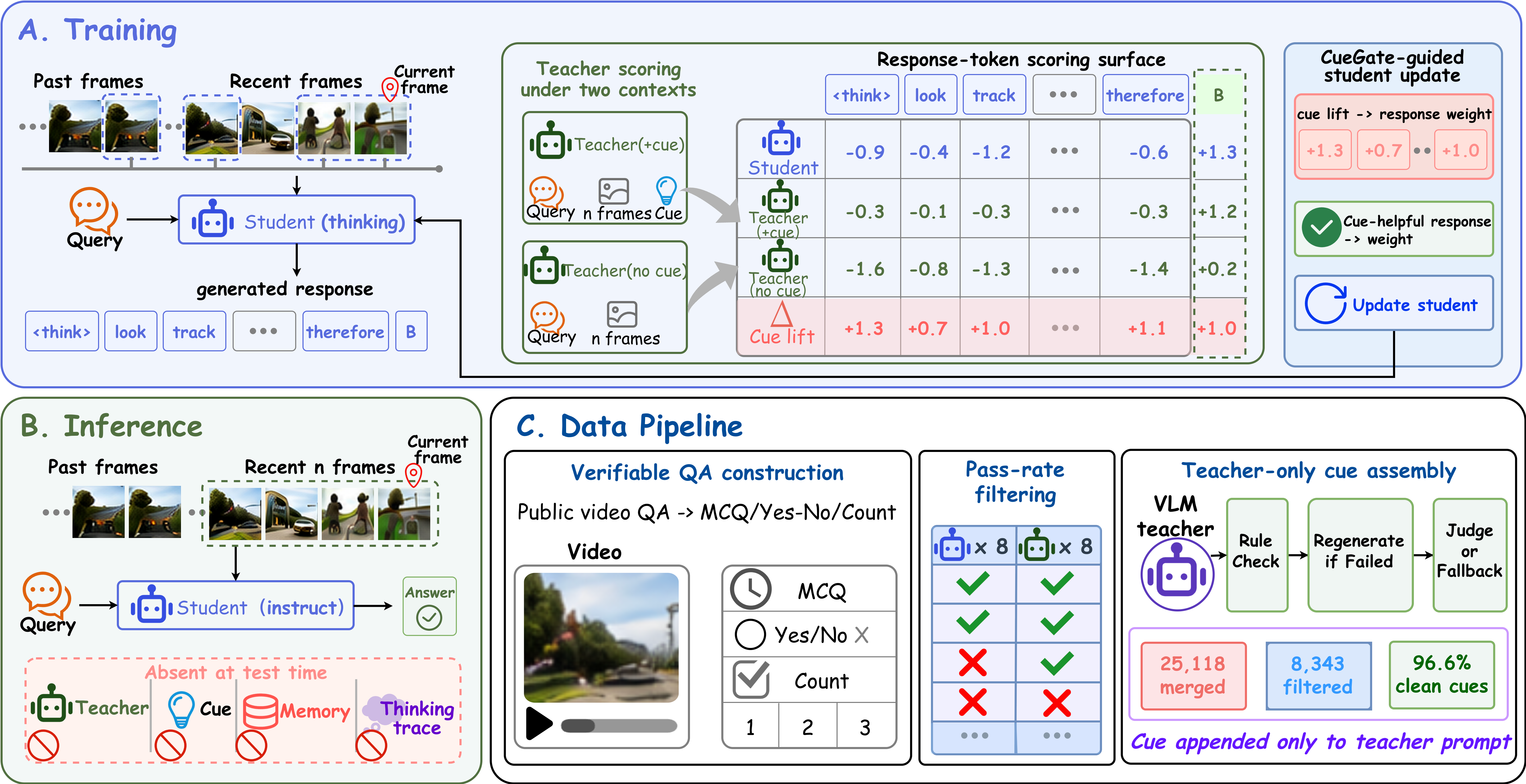}
\caption{StreamOPD pipeline with ST-CueGate. (a) During thinking-mode training, the student generates an on-policy response $y$ from a video sampled with the standard training pipeline. The cue-augmented teacher forward provides the distillation target $\tau^{+}$, while the no-cue forward provides a nested reference $\tau^{-}$ on the same frames and response. Their token-wise conditional log-likelihood ratio is aggregated into a response-level weight $w$ that gates the OPD advantage. (b) During instruct-mode inference, the student uses the recent-window protocol---frames only, with no cue, memory module, or generated thinking trace.}
\label{fig:method}
\end{figure*}

\paragraph{Problem Setup and Inference Protocol.}
A streaming VLM receives a question $q$ and a sequence of frames arriving over time; at the query moment it must answer from the frames seen so far. Following SimpleStream~\citep{shen2026simple}, we fix a \emph{memory-free recent-window} inference protocol: the model conditions only on the most recent frames within a small budget (a recent-4-frame window at $1$\,fps), with no memory bank, retrieval, or compression. This keeps inference cost and architecture constant, so any improvement is attributable to the model's weights rather than test-time machinery. Answers fall into three verifiable formats---multiple choice (MCQ), binary (Yes/No), and counting. Their rule-based reward $r\in\{0,1\}$ is used only by the GRPO diagnostic in Experiments; the main OPD objective uses teacher log-probabilities rather than task reward. The unified backbone supports both thinking and direct instruct modes; since only both-thinking training converges among the tested pairings, we train teacher and student in thinking mode and deploy the student in instruct mode.

\paragraph{On-Policy Distillation Backbone.}
Let $\mathcal{D}=\{q_i\}_{i=1}^{N}$ denote the training prompts (each paired with its video input), $\pi_\theta$ the student, and $\pi_\tau$ a frozen teacher. For each $q_i$, the student samples an on-policy response $y_i=(y_{i,1},\ldots,y_{i,T_i})\sim\pi_\theta(\cdot\mid q_i)$. On a valid response token $t$, define
\[
s_{i,t}=\log\pi_\theta(y_{i,t}\mid y_{i,<t},q_i),
\qquad
\tau_{i,t}=\log\pi_\tau(y_{i,t}\mid y_{i,<t},q_i).
\]
Standard OPD uses the sampled-token k1 reverse-KL signal
\begin{equation}
A_{i,t}^{\text{OPD}}=\tau_{i,t}-s_{i,t},
\label{eq:opd}
\end{equation}
treated as a stop-gradient advantage on the student's own rollout. Conceptually, all variants optimize the same token-level policy-gradient objective,
\begin{equation}
\mathcal{L}_{\mathrm{PG}}(\theta;A)
=-\mathbb{E}_{q_i,y_i}\!\left[
\frac{1}{T_i}\sum_{t=1}^{T_i}
\log\pi_\theta(y_{i,t}\mid y_{i,<t},q_i)\,
\mathrm{sg}\!\left[A_{i,t}\right]
\right],
\label{eq:pg-objective}
\end{equation}
implemented with clipped importance ratios ~\citep{schulman2017proximal} to the rollout policy. Standard OPD sets $A_{i,t}=A_{i,t}^{\mathrm{OPD}}$; the privilege variants below modify only this advantage. The teacher need not be larger: the same formulation supports on-policy self-distillation (OPSD)~\citep{zhao2026self} by using a frozen copy of the student's initial policy under a different conditioning context.

A practical subtlety is that teacher and student consume different multimodal prompts once teacher-side privilege is introduced, so their prompt lengths differ. We therefore extract teacher log-probabilities over the \emph{response tokens only} and align them to the student's generated tokens, decoupling teacher-side context from token-level supervision.

\paragraph{From Giving to Gating Teacher Privilege.}
Beyond ``how strong is the teacher,'' we ask ``what is the teacher given''---and how its predictive distribution changes under that condition. We call a teacher condition \emph{clip-grounded} when its content is generated from the same training clip rather than copied from an answer-side oracle; the stored answer is used only to reject potentially leaking cue candidates. This definition concerns training-time provenance: the cue text itself is teacher-only and is never available during causal recent-window inference. We first review conditions that only \emph{supply} privilege, then present \textbf{Spatio-Temporal CueGate (ST-CueGate)}, which computes a likelihood-based sensitivity proxy and uses it to reweight OPD. All variants share an identical backbone, dataset, and optimizer, so differences are attributable to the teacher condition alone. For the extrapolation baseline we also use $b_{i,t}$, the token log-probability under the frozen student initialization.

\paragraph{Ungrounded Baselines.}
\emph{Reward extrapolation} (ExOPD) tries to exceed the teacher by extrapolating its improvement over the base by a factor $\lambda\ge 1$,
\begin{equation}
A_{i,t}^{\text{ExOPD}}
=-\big[(s_{i,t}-b_{i,t})-\lambda(\tau_{i,t}-b_{i,t})\big],
\label{eq:exopd}
\end{equation}
with $\lambda=1$ recovering standard OPD. This baseline manipulates the learning signal without conditioning it on an observation available to the student. Additional ungrounded reweighting controls are deferred to the appendix.

\paragraph{Passive Cue Conditioning.}
ViCuR~\citep{tian2026vicur} treats visual cues as recoverable teacher privilege and additionally trains a student-side cue recovery module. To isolate teacher-side conditioning while keeping the student fixed, we use its cue-only variant, denoted \emph{ViCuR (cue-only)}: a short, automatically screened spatio-temporal cue is appended only to the teacher prompt. This baseline supplies the cue uniformly; ST-CueGate instead measures the cue-induced teacher shift on each student response and gates distillation by that contrast.

\begin{figure}[t]
\centering
\includegraphics[width=\columnwidth]{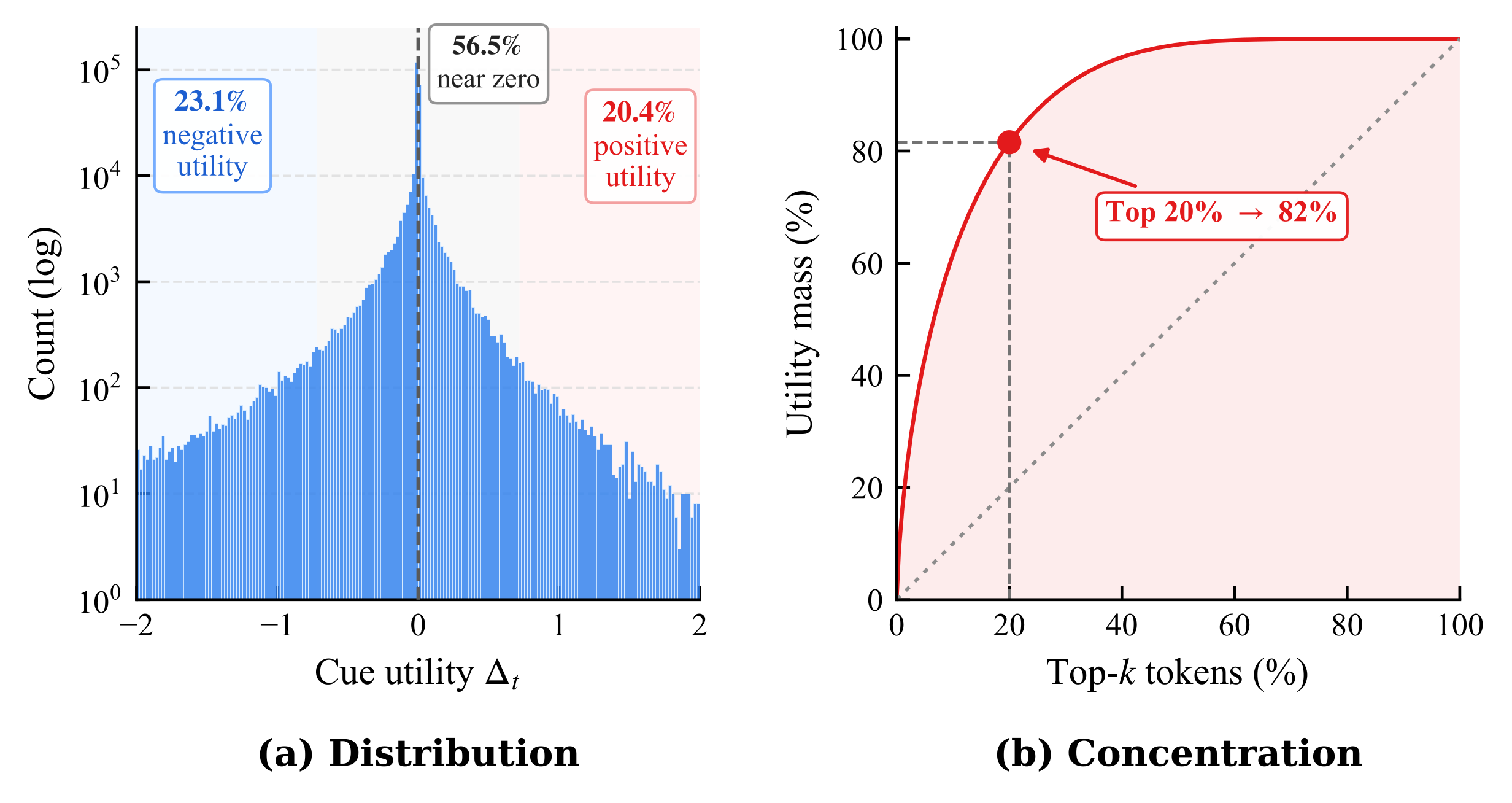}
\caption{Distribution and concentration of the cue-conditional log-likelihood ratio over 300,866 student-response tokens. (a) Most token scores are near zero, while the remaining distribution is heavy-tailed and includes both positive and negative shifts. (b) The top $20\%$ of positive-score tokens account for $82\%$ of the total positive mass.}
\label{fig:cue-delta-distribution}
\end{figure}

\paragraph{ST-CueGate: Nested Cue-Removal Gating.}
Passive conditioning assumes that the cue benefits every rollout equally. ST-CueGate instead measures its contribution on the student's own response (Fig.~\ref{fig:method}). Given an automatically accepted spatio-temporal cue $c_i$, the frozen teacher scores the same response under two conditioning contexts:
\begin{align}
\tau_{i,t}^{+}
&=\log\pi_\tau(y_{i,t}\mid y_{i,<t},q_i,c_i),\notag\\
\tau_{i,t}^{-}
&=\log\pi_\tau(y_{i,t}\mid y_{i,<t},q_i).
\label{eq:teacher-views}
\end{align}
The cue-augmented score $\tau_{i,t}^{+}$ is the distillation target used by both cue-only ViCuR and ST-CueGate, i.e., $\tau_{i,t}=\tau_{i,t}^{+}$ in Eq.~\ref{eq:opd}. The no-cue score $\tau_{i,t}^{-}$ is used only to form the token-level conditional contrast,
\begin{equation}
\begin{aligned}
\Delta_{i,t}
&=\tau_{i,t}^{+}-\tau_{i,t}^{-}\\
&=\log\frac{\pi_\tau(y_{i,t}\mid y_{i,<t},q_i,c_i)}
{\pi_\tau(y_{i,t}\mid y_{i,<t},q_i)}.
\end{aligned}
\label{eq:cuedelta}
\end{equation}
A positive $\Delta_{i,t}$ means that the cue increases the frozen teacher's confidence in the realized token $y_{i,t}$. This is a pointwise conditional log-likelihood ratio between two nested contexts that hold the teacher, frames, question, prefix, and token fixed. Empirically, this teacher-side contrast is highly non-uniform (Fig.~\ref{fig:cue-delta-distribution}): $56.5\%$ of tokens have near-zero scores ($|\Delta|\leq0.01$), $23.1\%$ have negative scores, and the top $20\%$ of positive-score tokens carry $82\%$ of the total positive mass. This concentration motivates measuring cue sensitivity before reweighting.

We aggregate the token-level contrast into a length-normalized response proxy,
\begin{equation}
\begin{aligned}
g_i
&=\frac{1}{T_i}\sum_{t=1}^{T_i}\Delta_{i,t}\\
&=\frac{1}{T_i}\log
\frac{\pi_\tau(y_i\mid q_i,c_i)}
{\pi_\tau(y_i\mid q_i)}.
\end{aligned}
\label{eq:cue-score}
\end{equation}
which summarizes the cue-induced shift in the teacher's likelihood of the same realized response. We then standardize it over a comparison group $\mathcal{G}(i)$. For $n{=}1$, $\mathcal{G}(i)$ is the minibatch; for $n{>}1$, it can be the sibling rollouts sharing the same prompt. Using population statistics,
\begin{align}
\mu_{\mathcal{G}}
&=\frac{1}{|\mathcal{G}|}\sum_{j\in\mathcal{G}(i)}g_j,\notag\\
\sigma_{\mathcal{G}}
&=\sqrt{\frac{1}{|\mathcal{G}|}
\sum_{j\in\mathcal{G}(i)}(g_j-\mu_{\mathcal{G}})^2},\notag\\
\widetilde{g}_i
&=\frac{g_i-\mu_{\mathcal{G}}}{\sigma_{\mathcal{G}}+\epsilon},\notag\\
w_i
&=\mathrm{sg}\!\left[
\mathrm{clip}\big(1+\alpha_{\mathrm{g}}\widetilde{g}_i,
w_{\min},w_{\max}\big)\right],\notag\\
A_{i,t}^{\mathrm{ST\text{-}CueGate}}
&=w_i\big(\tau_{i,t}^{+}-s_{i,t}\big).
\label{eq:cuegate}
\end{align}
Responses with an above-group-mean score receive $w_i>1$, while below-mean responses receive $w_i<1$. This ranking is relative, not absolute: if every $g_i$ in a group is negative, a less-negative response can still be up-weighted. Moreover, although $\Delta_{i,t}$ is computed per token, ST-CueGate applies one scalar $w_i$ to every OPD token advantage in the response; it therefore does not localize which tokens benefit from the cue. Averaging removes first-order scaling with token count but remains sensitive to response composition, formatting tokens, and teacher calibration. We consequently interpret $g_i$ as a likelihood-based relative cue-sensitivity proxy rather than faithful utility attribution. We use $\alpha_{\mathrm{g}}=0.5$ and $[w_{\min},w_{\max}]=[0,2]$; a singleton comparison group is assigned the neutral gate $w_i=1$. Setting $\alpha_{\mathrm{g}}=0$ yields $w_i\equiv1$ and therefore the same distillation gradient as cue-only ViCuR. The no-cue forward reuses the student's decoded frames and the same teacher pool, adding no GPU allocation; it affects training only through $w_i$.

\begin{figure}[t]
\centering
\includegraphics[width=0.9\columnwidth]{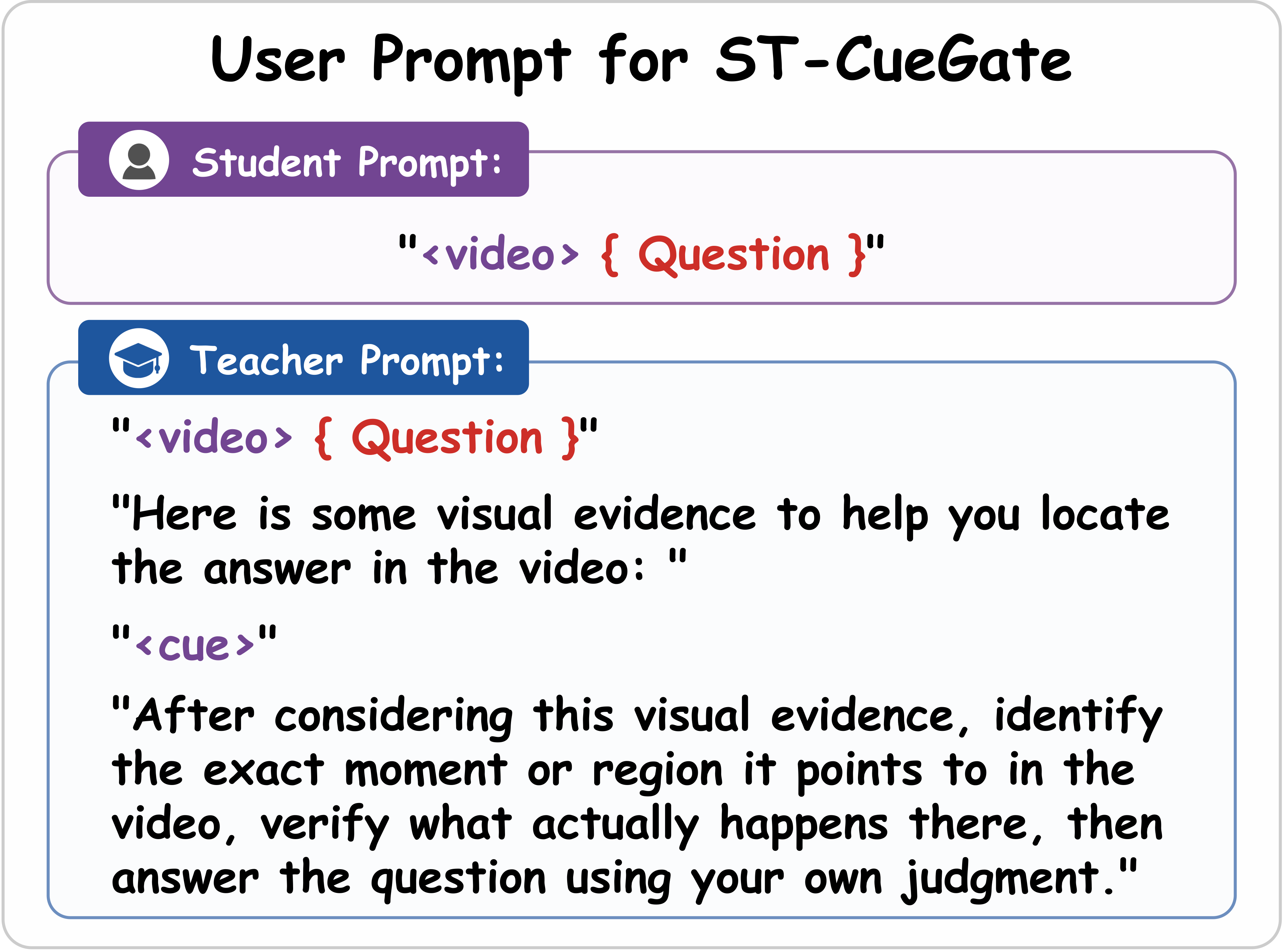}
\caption{Prompt formats for ST-CueGate. The student receives the original video question, while the teacher additionally receives an automatically screened spatio-temporal cue. ST-CueGate also scores the same response under the original no-cue prompt.}
\label{fig:prompt_templates}
\end{figure}

\paragraph{Training Data and Visual Cues.}
We construct an approximately 25k-item training set of verifiable video questions in MCQ, binary, and counting formats from public instruction data~\citep{zhang2024llava,yuan2025tarsier2}. A stronger frozen VLM processes each complete short training clip and generates a spatio-temporal pointer after answer options are removed from its input. Rule-based screening, regeneration, and a deterministic judging pass by the same generator model automatically accept ${\sim}96.6\%$ of cues; rejected samples fall back to the original teacher prompt. To complement this correlated self-check, two annotators jointly review a random sample of 300 accepted cues and flag $3$ direct-answer leaks, $4$ indirect semantic leaks, and $6$ visually unsupported or incorrectly grounded pointers, i.e.\ a residual leakage rate of $7/300$ among accepted cues. The cue is added only to the teacher through the explicit instruction template in Fig.~\ref{fig:prompt_templates}; the student prompt and inference protocol remain unchanged. Full construction, model choice, and audit details are provided in the Supplementary Material.

\section{Experiments}

\paragraph{Setup.}
We use a Qwen3.5-4B student and Qwen3.5-9B teacher. Both train in thinking mode with the OPD objective (Eq.~\ref{eq:opd}) and deploy in instruct mode. Each in-house configuration is trained with three independent seeds; within a run we select a single model on a held-out validation aggregate and evaluate that same model on all four benchmarks, and reported entries are means over the three runs. StreamingBench~\citep{lin2026streamingbench} and OVO-Bench~\citep{niu2025ovo} use recent-4 frames at 1\,fps; Video-MME~\citep{fu2025video} and LongVideoBench~\citep{wu2024longvideobench} use their standard protocols with at most 32 frames. All evaluations use greedy decoding, and OVO is reported with and without HLD. Full training, implementation, and statistical details are provided in the Supplementary Material.

\paragraph{StreamOPD Core Recipe Under Fixed Recent-Window Inference.}
Table~\ref{tab:main} shows that standard OPD closes most of the 4B--9B gap under unchanged inference: StreamingBench rises from $77.9\%$ to $83.9\%$, and OVO-Bench excluding HLD improves by $9.1$ points. The gains span both real-time perception and backward tracing, but they are unevenly distributed across subtasks: STU ($+20.7$) and OCR ($+13.4$) improve most, the remaining real-time subtasks gain between $3.0$ and $9.2$ points, and backward tracing improves on both EPM ($+9.1$) and ASI ($+7.4$). LongVideoBench improves modestly and Video-MME slips slightly below the base model. HLD moves in the opposite direction, indicating that dense teacher supervision strengthens answerable-task competence but can weaken abstention; we therefore report it separately following SimpleStream~\citep{shen2026simple}.

\begin{table*}[t!]
\centering
\newcommand{\best}[1]{\textbf{#1}}
\setlength{\tabcolsep}{3pt}
\small
\begin{tabular}{lc|c|cccccc|c|ccc|c|c}
\toprule
\multirow{3}{*}{Model} & \multirow{3}{*}{\#Frames}
& \multirow{3}{*}{\shortstack{Streaming\\Bench}}
& \multicolumn{12}{c}{\textit{OVO-Bench}} \\
\cmidrule(lr){4-15}
& &
& \multicolumn{7}{c|}{\textit{Real-Time Visual Perception}}
& \multicolumn{4}{c|}{\textit{Backward Tracing}}
& \multirow{2}{*}{Avg.} \\
\cmidrule(lr){4-10} \cmidrule(lr){11-14}
& & & OCR & ACR & ATR & STU & FPD & OJR & Avg.
& EPM & ASI & HLD & Avg. & \\
\midrule
Human
& -- & 91.46
& 94.0 & 92.6 & 94.8 & 92.7 & 91.1 & 94.0 & 93.2
& 92.6 & 93.0 & 91.4 & 92.3
& 92.77 \\
\midrule
\multicolumn{15}{c}{\textit{Offline Video LLMs}} \\
\midrule
Qwen2.5-VL-7B~\citep{bai2025qwen25vltechnicalreport}
& 1\,fps & 73.31
& 67.8 & 55.1 & 67.2 & 42.1 & 66.3 & 60.9 & 59.9
& 51.5 & 58.8 & 23.7 & 44.7
& 52.28 \\
LLaVA-OneVision-7B~\citep{li2024llava}
& 32 & 71.12
& 66.4 & 57.8 & 73.3 & 53.4 & 71.3 & 62.0 & 64.0
& 54.2 & 55.4 & 21.5 & 43.7
& 53.85 \\
LLaVA-Video-7B~\citep{zhang2024llava}
& 64 & --
& 69.1 & 58.7 & 68.8 & 49.4 & 74.3 & 59.8 & 63.5
& \underline{56.2} & 57.4 &  7.5 & 40.4
& 51.95 \\
LongVU-7B~\citep{shen2024longvu}
& 1\,fps & --
& 55.7 & 49.5 & 59.5 & 48.3 & 68.3 & 63.0 & 57.4
& 43.1 & \best{66.2} &  9.1 & 39.5
& 48.45 \\
\midrule
\multicolumn{15}{c}{\textit{Online / Streaming Video LLMs}} \\
\midrule
Dispider-7B~\citep{qian2025dispider}
& 1\,fps & 67.63
& 57.7 & 49.5 & 62.1 & 44.9 & 61.4 & 51.6 & 54.6
& 48.5 & 55.4 &  4.3 & 36.1
& 45.35 \\
TimeChat-Online-7B~\citep{yao2025timechat}
& 1\,fps & 75.28
& 75.2 & 46.8 & 70.7 & 47.8 & 69.3 & 61.4 & 61.9
& 55.9 & 59.5 &  9.7 & 41.7
& 51.80 \\
StreamForest-7B~\citep{zeng2026streamforest}
& 1\,fps & 77.26
& 68.5 & 53.2 & 71.6 & 47.8 & 65.4 & 60.9 & 61.2
& \best{58.9} & 64.9 & 32.3 & 52.0
& 56.60 \\
HERMES-7B$^\dagger$~\citep{zhang2026hermes}
& 1\,fps & 79.44
& 85.2 & 64.2 & 71.6 & 53.4 & 74.3 & 65.2 & 69.0
& 48.5 & 62.2 & 37.6 & 49.4
& 59.20 \\
\midrule
\multicolumn{15}{c}{\textit{Post-Trained Models and References (\textsc{StreamOPD}, Ours)}} \\
\midrule
Qwen3.5-9B (teacher)
& 4 & \underline{84.15}
& \underline{95.3} & 78.0 & \best{87.9} & 70.8 & \underline{74.3} & \best{85.9} & \underline{82.0}
& 54.9 & 59.5 & 47.3 & 53.9
& \underline{67.95} \\
Qwen3.5-4B (student)
& 4 & 77.87
& 81.9 & 67.9 & 78.5 & 52.3 & 68.3 & 75.0 & 70.6
& 47.1 & 51.4 & \underline{47.9} & 48.8
& 59.71 \\
\ \ + OPD
& 4 & 83.91
& \underline{95.3} & 77.1 & \underline{84.5} & \best{73.0} & 71.3 & 82.1 & 80.5
& 56.2 & 58.8 & 38.7 & 51.2
& 65.89 \\
\ \ + \textbf{ST-CueGate} (OPD)
& 4 & \best{84.55}
& \best{98.0} & \best{80.7} & 83.6 & \underline{71.9} & \best{76.2} & \underline{85.3} & \best{82.6}
& \best{58.9} & 63.5 & 45.7 & \best{56.0}
& \best{69.34} \\
\ \ + \textbf{ST-CueGate} (OPSD)
& 4 & 83.35
& 93.3 & 74.3 & 82.8 & 71.4 & \underline{74.3} & 79.4 & 79.2
& 54.2 & 56.8 & \best{57.0} & \best{56.0}
& 67.60 \\
\bottomrule
\end{tabular}
\caption{Main results on OVO-Bench~\citep{niu2025ovo} and StreamingBench~\citep{lin2026streamingbench}. We report OVO-Bench per-task accuracy under its Real-Time and Backward tracks; ``Avg.'' is their mean and ``--'' denotes unreported results. All \textsc{StreamOPD} models are evaluated in instruct mode with a recent-4-frame window. ST-CueGate (OPD) uses the 9B teacher, while ST-CueGate (OPSD)~\citep{zhao2026self} uses the frozen 4B initial policy; $\dagger$ denotes Qwen2.5-VL-7B + HERMES (4K tokens).}
\label{tab:main}
\end{table*}

\paragraph{Training--Deployment Mode Decoupling and Stability.}
Because the backbone is a unified thinking/instruct model, we can decouple the training configuration from the response mode used at deployment. Among the tested pairings, both-instruct and teacher-thinking/student-instruct training collapse, whereas both-thinking training converges. This comparison jointly changes response length, policy distributions, and teacher--student alignment; it therefore establishes mode-dependent instability in our setting but not response length as its sole cause. We train OPD with both policies in thinking mode and evaluate the same distilled policy in both decoding modes. Instruct-mode evaluation performs better while preserving concise option-format outputs, matching the benchmark's direct-answer format. We use this empirically selected train-thinking / deploy-instruct pairing throughout.

\paragraph{GRPO Format Drift: A Response-Format Diagnostic.}
Without a teacher, GRPO drifts from concise answers to long rationales that place the answer at the end. It therefore appears strong under the reward's last-letter parser but fails under the deployed first-answer parser: the same checkpoint scores $82.4\%$ under the former and only $35.1\%$ under the latter, while its median response length grows more than tenfold over training. Distilled models remain concise and parser-invariant, indicating that dense supervision anchors response format as well as answer content; detailed diagnostics are provided in the Supplementary Material.

\paragraph{Teacher-Privilege Extension: ST-CueGate.}
With the core recipe fixed, we next ask whether \emph{the way the teacher is conditioned} can improve axes left unresolved by standard OPD. Table~\ref{tab:privilege} compares teacher-side conditioning and gating designs on the same 25k data pool. Ungrounded extrapolation and passive cue conditioning do not consistently improve the four benchmarks, motivating a cue-specific gate. Additional controls are reported in the Supplementary Material.

ST-CueGate improves all four benchmarks over both $n{=}1$ and rollout-matched $n{=}4$ OPD. Against the latter, the gains are $+0.7$, $+1.8$, $+0.8$, and $+0.2$ points, raising the cross-benchmark mean from $69.81$ to $70.69$. The largest improvement occurs on OVO-Bench, suggesting that cue-specific reweighting is especially useful when answers depend on locating temporal evidence. ST-CueGate also exceeds cue-only ViCuR on all four benchmarks, showing that the gain comes from selectively weighting cue-conditioned supervision rather than from supplying the cue alone. Reading the table against the untrained student in its first row exposes a cost that the streaming numbers alone hide: every other variant trades general-video ability for streaming gains and ends up below the base model on Video-MME, whereas ST-CueGate is the only configuration that stays above the base on all four benchmarks.

The per-subtask breakdown in Table~\ref{tab:main} shows that gating does more than move the student closer to its teacher. ST-CueGate exceeds the 9B teacher on six of the nine OVO-Bench subtasks---OCR, ACR, STU, FPD, EPM, and ASI, by $1.1$ to $4.0$ points---as well as on the OVO-Bench macro ($69.34$ versus $67.95$) and on StreamingBench ($84.55$ versus $84.15$). The residual teacher advantage is concentrated in ATR, OJR, and HLD. A 4B student can therefore overtake the 9B teacher that supervises it on most streaming subtasks, which is consistent with on-policy supervision shaping the student along its own trajectory distribution rather than transferring a fixed teacher ceiling. The exception is again abstention. The post-trained 4B model also compares favorably with the larger streaming systems in Table~\ref{tab:main}: it exceeds HERMES-7B, the strongest listed streaming design, by $5.1$ points on StreamingBench and $10.1$ points on the OVO-Bench macro, while operating on a strictly memory-free four-frame context.

To test portability beyond larger-teacher OPD, we also instantiate the $n{=}4$ ST-CueGate configuration within OPSD using a frozen 4B same-backbone teacher. Without a larger model, this variant reaches $83.35\%$ on StreamingBench and $67.35\%$ on OVO-Bench excluding HLD, so ST-CueGate is compatible with self-distillation rather than being tied to teacher scale. It also attains the highest HLD score in Table~\ref{tab:main} at $57.0\%$, above both the untrained student ($47.9\%$) and the 9B teacher ($47.3\%$). The abstention loss seen under larger-teacher distillation is therefore not intrinsic to the recipe, and does not appear in our same-backbone instantiation.

The unifying pattern is \emph{clip-grounded provenance with a cue-specific reference}. Injecting a signal unanchored to the training video does not help, while grounded privilege remains insufficient when supplied passively. ST-CueGate instead holds the frames and realized trajectory fixed while removing only the cue; this reference performs better than a shuffled-video negative view.

\begin{table*}[t]
\centering
\small
\setlength{\tabcolsep}{5pt}
\begin{tabular}{@{}llccccc@{}}
\toprule
\textbf{Method} &
\textbf{Grounded} &
\textbf{StreamingBench} &
\textbf{OVO-Bench (excl.\ HLD)} &
\textbf{Video-MME} &
\textbf{LongVideoBench} &
\textbf{Avg.} \\
\midrule
Qwen3.5-4B (student)             & --                        & 77.87 & 59.94 & 64.22 & 57.74 & 64.94 \\
\midrule
OPD ($n{=}1$)                    & --                        & 83.91 & 69.02 & 63.33 & 59.84 & 69.03 \\
OPD ($n{=}4$)                    & --                        & 83.83 & 70.09 & 64.07 & 61.26 & 69.81 \\
ExOPD ($n{=}1$)                  & \ding{55}                 & 84.29 & 68.38 & 63.78 & 60.36 & 69.20 \\
ViCuR (cue-only, $n{=}1$)        & \ding{51}                 & 84.27 & 67.98 & 60.56 & 54.00 & 66.70 \\
ViCuR (cue-only, $n{=}4$)        & \ding{51}                 & 84.49 & 67.86 & 64.11 & 60.73 & 69.30 \\
V-Zero ($n{=}4$)                 & \ding{51}                 & 83.99 & 67.71 & 61.67 & 60.06 & 68.36 \\
\textbf{ST-CueGate} ($n{=}4$)    & \ding{51}$^{\ast}$        & \textbf{84.55} & \textbf{71.93} & \textbf{64.85} & \textbf{61.41} & \textbf{70.69} \\
\bottomrule
\end{tabular}
\caption{Ablation of teacher conditioning and cue gating on the 25k data pool. The first row is the untrained student. ``ViCuR (cue-only)'' retains only ViCuR's teacher-side visual cue, without its student recovery module~\citep{tian2026vicur}; our video implementation of V-Zero~\citep{sun2026v} uses a temporally shuffled video as the negative evidence view. Grounded denotes a teacher-side signal derived from the same underlying training clip; $^{\ast}$ denotes nested cue-removal gating. The OVO-Bench column excludes HLD; raw HLD and the full macro are reported separately in Table~\ref{tab:main}. Avg.\ is the unweighted mean over the four benchmarks.}
\label{tab:privilege}
\end{table*}

\begin{table}[t]
\centering
\small
\setlength{\tabcolsep}{1.5pt}
\begin{tabular}{@{}lccccc@{}}
\toprule
\textbf{Gate Setting} &
\shortstack{\textbf{Streaming}\\\textbf{Bench}} &
\shortstack{\textbf{OVO-Bench}\\\textbf{(excl.\ HLD)}} &
\shortstack{\textbf{Video-}\\\textbf{MME}} &
\shortstack{\textbf{LongVideo}\\\textbf{Bench}} &
\textbf{Avg.} \\
\midrule
$n{=}1$, batch, seq. & 84.23 & 68.93 & \textbf{65.11} & 59.69 & 69.49 \\
$n{=}4$, uid, seq.   & \textbf{84.55} & \textbf{71.93} & 64.85 & \textbf{61.41} & \textbf{70.69} \\
$n{=}4$, uid, token  & 84.23 & 68.72 & 63.52 & 59.99 & 69.12 \\
$n{=}8$, uid, seq.   & 84.11 & 68.81 & 64.41 & 60.66 & 69.50 \\
\bottomrule
\end{tabular}
\caption{Ablation of ST-CueGate rollout grouping and gate granularity on the 25k data pool. ``seq.'' denotes one response-level gate; ``token'' denotes token-wise gates. Avg.\ is the unweighted mean over the four benchmarks.}
\label{tab:ablation-group}
\end{table}

Table~\ref{tab:ablation-group} shows that response-level $n{=}4$ gating performs best on StreamingBench, OVO-Bench, and LongVideoBench, whereas $n{=}1$ slightly favors Video-MME. Token-wise gating loses most on OVO-Bench ($-3.2$) and LongVideoBench ($-1.4$), suggesting that individual token contrasts are too noisy to serve as independent weights; aggregating them at the response level provides a more stable signal.

\begin{table}[t]
\centering
\small
\setlength{\tabcolsep}{1.5pt}
\begin{tabular}{@{}lccccc@{}}
\toprule
\textbf{$\alpha_{\mathrm{g}}$} &
\shortstack{\textbf{Streaming}\\\textbf{Bench}} &
\shortstack{\textbf{OVO-Bench}\\\textbf{(excl.\ HLD)}} &
\shortstack{\textbf{Video-}\\\textbf{MME}} &
\shortstack{\textbf{LongVideo}\\\textbf{Bench}} &
\textbf{Avg.} \\
\midrule
0.25 & 84.19 & 69.09 & 64.19 & 60.73 & 69.55 \\
0.50 & \textbf{84.55} & \textbf{71.93} & \textbf{64.85} & \textbf{61.41} & \textbf{70.69} \\
1.00 & 84.39 & 68.97 & 63.22 & 59.76 & 69.09 \\
\bottomrule
\end{tabular}
\caption{Ablation of the ST-CueGate strength $\alpha_{\mathrm{g}}$ with fixed $n{=}4$, UID sibling normalization, $\gamma{=}1$, and gate range $[0,2]$. Avg.\ is the unweighted mean over the four benchmarks.}
\label{tab:ablation-alpha}
\end{table}

Table~\ref{tab:ablation-alpha} varies the gate strength: $\alpha_{\mathrm{g}}=0.5$ gives the highest result on all four benchmarks. Increasing the strength to $1.0$ consistently hurts performance, indicating that the likelihood contrast is most useful as a moderate ranking signal rather than an aggressively amplified weight.

\begin{table}[t]
\centering
\small
\setlength{\tabcolsep}{1.5pt}
\begin{tabular}{@{}lccccc@{}}
\toprule
\textbf{Teacher} &
\shortstack{\textbf{Streaming}\\\textbf{Bench}} &
\shortstack{\textbf{OVO-Bench}\\\textbf{(excl.\ HLD)}} &
\shortstack{\textbf{Video-}\\\textbf{MME}} &
\shortstack{\textbf{LongVideo}\\\textbf{Bench}} &
\textbf{Avg.} \\
\midrule
\textbf{Qwen3.5-9B} & \textbf{84.55} & \textbf{71.93} & \textbf{64.85} & \textbf{61.41} & \textbf{70.69} \\
Qwen3.5-27B          & 83.59 & 67.52 & 61.04 & 54.38 & 66.63 \\
\bottomrule
\end{tabular}
\caption{Ablation of teacher model size for ST-CueGate with a fixed Qwen3.5-4B student. Avg.\ is the unweighted mean over the four benchmarks.}
\label{tab:ablation-teacher}
\end{table}

Table~\ref{tab:ablation-teacher} varies teacher scale with the student fixed. The larger teacher is worse on all four benchmarks, with the smallest deficit on StreamingBench ($-1.0$) and the largest on LongVideoBench ($-7.0$). This run reuses the 9B optimization and gate settings without re-tuning, so it should be read as evidence that teacher scale alone does not guarantee better on-policy supervision, not as a tuned upper bound for a 27B teacher: a closer-capacity teacher may provide token distributions better aligned with the student's trajectories, while a larger teacher may require different optimization or gating calibration.

\section{Related Work}

\paragraph{Streaming Video Understanding.} Unlike offline video MLLMs that process complete clips~\citep{qwen37,bai2023qwen,wang2024qwen2,bai2025qwen25vltechnicalreport,zhang2024llava}, streaming models operate on a causal video prefix. Prior systems rely on memory, retrieval, or KV-cache compression~\citep{chen2024videollm,zhang2024flash,qian2025dispider,yao2025timechat,zeng2026streamforest,zhang2026hermes,xiao2024efficient}, with StreamingBench and OVO-Bench evaluating real-time and historical reasoning~\citep{lin2026streamingbench,niu2025ovo}. SimpleStream~\citep{shen2026simple} shows that a recent-window baseline can match heavier designs, whereas ThinkStream~\citep{liu2026thinking} combines memory with online reasoning.

\paragraph{On-Policy Distillation and Teacher Privilege.} Classical distillation uses teacher targets on fixed data~\citep{hinton2015distilling}, whereas OPD scores student-generated trajectories~\citep{agarwal2024policy,gu2024minillm} and has been extended to text reasoning~\citep{zhao2026self}, multimodal pre-alignment~\citep{wang2026beyond}, and temporal video grounding~\citep{li2026video}. Privileged-information methods provide training-only side information~\citep{vapnik2015learning,lopez2015unifying}, with recent work exploring visual-cue privilege and contrastive evidence gating~\citep{tian2026vicur,sun2026v}. V-Zero contrasts question-relevant evidence with a negative visual view to gate answer-label-free OPD. ST-CueGate shares the response-level contrastive reweighting structure, but uses a nested cue-removal reference that holds the frames, question, and realized student trajectory fixed and changes only the teacher-side spatio-temporal cue.

\section{Conclusion}
Under a fixed memory-free recent-window protocol, we isolate post-training from added inference machinery. Our study exposes two empirical failure modes: teacher-free GRPO drifts toward verbose, deployment-incompatible responses, while the tested OPD configurations involving instruct-mode training collapse. Together with the 25k verifiable-data pipeline and thinking-train/instruct-infer configuration, these diagnostics form the StreamOPD core recipe. Standard OPD places the 4B student within $0.3$ points of the 9B teacher on StreamingBench and substantially improves OVO-Bench under unchanged inference. As a teacher-privilege extension, ST-CueGate uses a nested cue-removal likelihood ratio to form a group-relative response proxy while keeping the student architecture fixed, yielding additional gains on OVO-Bench and Video-MME. Its successful same-backbone OPSD instantiation further shows that the mechanism is not tied to a larger teacher. We hope \textsc{StreamOPD} serves as a transparent empirical foundation and open-source reference for reproducible streaming-video post-training.

\bibliography{aaai2027}

@misc{qwen35blog,
    title = {Qwen3.5: Accelerating Productivity with Native Multimodal Agents},
    url = {https://qwen.ai/blog?id=qwen3.5},
    author = {Qwen Team},
    month = {February},
    year = {2026}
}

@misc{qwen37,
    title = {{Qwen3.7}: The Agent Frontier},
    url = {https://qwen.ai/blog?id=qwen3.7},
    author = {{Qwen Team}},
    month = {May},
    year = {2026}
}

@misc{bai2025qwen25vltechnicalreport,
      title={Qwen2.5-VL Technical Report}, 
      author={Shuai Bai and Keqin Chen and Xuejing Liu and Jialin Wang and Wenbin Ge and Sibo Song and Kai Dang and Peng Wang and Shijie Wang and Jun Tang and Humen Zhong and Yuanzhi Zhu and Mingkun Yang and Zhaohai Li and Jianqiang Wan and Pengfei Wang and Wei Ding and Zheren Fu and Yiheng Xu and Jiabo Ye and Xi Zhang and Tianbao Xie and Zesen Cheng and Hang Zhang and Zhibo Yang and Haiyang Xu and Junyang Lin},
      year={2025},
      eprint={2502.13923},
      archivePrefix={arXiv},
      primaryClass={cs.CV},
      url={https://arxiv.org/abs/2502.13923}, 
}

@article{wang2024qwen2,
  title={Qwen2-vl: Enhancing vision-language model's perception of the world at any resolution},
  author={Wang, Peng and Bai, Shuai and Tan, Sinan and Wang, Shijie and Fan, Zhihao and Bai, Jinze and Chen, Keqin and Liu, Xuejing and Wang, Jialin and Ge, Wenbin and others},
  journal={arXiv preprint arXiv:2409.12191},
  year={2024}
}

@article{bai2023qwen,
  title={Qwen-vl: A versatile vision-language model for understanding, localization},
  author={Bai, Jinze and Bai, Shuai and Yang, Shusheng and Wang, Shijie and Tan, Sinan and Wang, Peng and Lin, Junyang and Zhou, Chang and Zhou, Jingren},
  journal={Text Reading, and Beyond},
  volume={2},
  number={1},
  pages={1},
  year={2023}
}

@inproceedings{lin2026streamingbench,
  title={Streamingbench: Assessing the gap for mllms to achieve streaming video understanding},
  author={Lin, Junming and Fang, Zheng and Chen, Chi and Cheng, Haoxuan and Wan, Zihao and Luo, Fuwen and Wang, Ziyue and Li, Peng and Liu, Yang and Sun, Maosong},
  booktitle={ICASSP 2026-2026 IEEE International Conference on Acoustics, Speech and Signal Processing (ICASSP)},
  pages={12147--12151},
  year={2026},
  organization={IEEE}
}

@inproceedings{niu2025ovo,
  title={Ovo-bench: How far is your video-llms from real-world online video understanding?},
  author={Niu, Junbo and Li, Yifei and Miao, Ziyang and Ge, Chunjiang and Zhou, Yuanhang and He, Qihao and Dong, Xiaoyi and Duan, Haodong and Ding, Shuangrui and Qian, Rui and others},
  booktitle={Proceedings of the Computer Vision and Pattern Recognition Conference},
  pages={18902--18913},
  year={2025}
}

@inproceedings{fu2025video,
  title={Video-mme: The first-ever comprehensive evaluation benchmark of multi-modal llms in video analysis},
  author={Fu, Chaoyou and Dai, Yuhan and Luo, Yongdong and Li, Lei and Ren, Shuhuai and Zhang, Renrui and Wang, Zihan and Zhou, Chenyu and Shen, Yunhang and Zhang, Mengdan and others},
  booktitle={Proceedings of the IEEE/CVF conference on computer vision and pattern recognition},
  pages={24108--24118},
  year={2025}
}

@article{wu2024longvideobench,
  title={Longvideobench: A benchmark for long-context interleaved video-language understanding},
  author={Wu, Haoning and Li, Dongxu and Chen, Bei and Li, Junnan},
  journal={Advances in Neural Information Processing Systems},
  volume={37},
  pages={28828--28857},
  year={2024}
}

@article{zhang2024llava,
  title={Llava-video: Video instruction tuning with synthetic data},
  author={Zhang, Yuanhan and Wu, Jinming and Li, Wei and Li, Bo and Ma, Zejun and Liu, Ziwei and Li, Chunyuan},
  journal={arXiv preprint arXiv:2410.02713},
  year={2024}
}

@article{yuan2025tarsier2,
  title={Tarsier2: Advancing large vision-language models from detailed video description to comprehensive video understanding},
  author={Yuan, Liping and Wang, Jiawei and Sun, Haomiao and Zhang, Yuchen and Lin, Yuan},
  journal={arXiv preprint arXiv:2501.07888},
  year={2025}
}

@article{hinton2015distilling,
  title={Distilling the knowledge in a neural network},
  author={Hinton, Geoffrey and Vinyals, Oriol and Dean, Jeff},
  journal={arXiv preprint arXiv:1503.02531},
  year={2015}
}

@inproceedings{agarwal2024policy,
  title={On-policy distillation of language models: Learning from self-generated mistakes},
  author={Agarwal, Rishabh and Vieillard, Nino and Zhou, Yongchao and Stanczyk, Piotr and Ramos Garea, Sabela and Geist, Matthieu and Bachem, Olivier},
  booktitle={International Conference on Learning Representations},
  volume={2024},
  pages={21246--21263},
  year={2024}
}

@inproceedings{gu2024minillm,
  title={Minillm: Knowledge distillation of large language models},
  author={Gu, Yuxian and Dong, Li and Wei, Furu and Huang, Minlie},
  booktitle={International Conference on Learning Representations},
  volume={2024},
  pages={32694--32717},
  year={2024}
}

@article{shao2024deepseekmath,
  title={Deepseekmath: Pushing the limits of mathematical reasoning in open language models},
  author={Shao, Zhihong and Wang, Peiyi and Zhu, Qihao and Xu, Runxin and Song, Junxiao and Bi, Xiao and Zhang, Haowei and Zhang, Mingchuan and Li, YK and Wu, Yang and others},
  journal={arXiv preprint arXiv:2402.03300},
  year={2024}
}

@article{guo2025deepseek,
  title={Deepseek-r1: Incentivizing reasoning capability in llms via reinforcement learning},
  author={Guo, Daya and Yang, Dejian and Zhang, Haowei and Song, Junxiao and Wang, Peiyi and Zhu, Qihao and Xu, Runxin and Zhang, Ruoyu and Ma, Shirong and Bi, Xiao and others},
  journal={arXiv preprint arXiv:2501.12948},
  year={2025}
}

@article{schulman2017proximal,
  title={Proximal policy optimization algorithms},
  author={Schulman, John and Wolski, Filip and Dhariwal, Prafulla and Radford, Alec and Klimov, Oleg},
  journal={arXiv preprint arXiv:1707.06347},
  year={2017}
}

@article{vapnik2015learning,
  title={Learning using privileged information: similarity control and knowledge transfer},
  author={Vapnik, Vladimir and Izmailov, Rauf},
  journal={The Journal of Machine Learning Research},
  volume={16},
  number={1},
  pages={2023--2049},
  year={2015},
  publisher={JMLR. org}
}

@article{lopez2015unifying,
  title={Unifying distillation and privileged information},
  author={Lopez-Paz, David and Bottou, L{\'e}on and Sch{\"o}lkopf, Bernhard and Vapnik, Vladimir},
  journal={arXiv preprint arXiv:1511.03643},
  year={2015}
}

@inproceedings{sheng2025hybridflow,
  title={Hybridflow: A flexible and efficient rlhf framework},
  author={Sheng, Guangming and Zhang, Chi and Ye, Zilingfeng and Wu, Xibin and Zhang, Wang and Zhang, Ru and Peng, Yanghua and Lin, Haibin and Wu, Chuan},
  booktitle={Proceedings of the Twentieth European Conference on Computer Systems},
  pages={1279--1297},
  year={2025}
}

@inproceedings{kwon2023efficient,
  title={Efficient memory management for large language model serving with pagedattention},
  author={Kwon, Woosuk and Li, Zhuohan and Zhuang, Siyuan and Sheng, Ying and Zheng, Lianmin and Yu, Cody Hao and Gonzalez, Joseph and Zhang, Hao and Stoica, Ion},
  booktitle={Proceedings of the 29th symposium on operating systems principles},
  pages={611--626},
  year={2023}
}

@inproceedings{xiao2024efficient,
  title={Efficient streaming language models with attention sinks},
  author={Xiao, Guangxuan and Tian, Yuandong and Chen, Beidi and Han, Song and Lewis, Mike},
  booktitle={International Conference on Learning Representations},
  volume={2024},
  pages={21875--21895},
  year={2024}
}

@inproceedings{chen2024videollm,
  title={Videollm-online: Online video large language model for streaming video},
  author={Chen, Joya and Lv, Zhaoyang and Wu, Shiwei and Lin, Kevin Qinghong and Song, Chenan and Gao, Difei and Liu, Jia-Wei and Gao, Ziteng and Mao, Dongxing and Shou, Mike Zheng},
  booktitle={Proceedings of the IEEE/CVF Conference on Computer Vision and Pattern Recognition},
  pages={18407--18418},
  year={2024}
}

@article{zhang2024flash,
  title={Flash-vstream: Memory-based real-time understanding for long video streams},
  author={Zhang, Haoji and Wang, Yiqin and Tang, Yansong and Liu, Yong and Feng, Jiashi and Dai, Jifeng and Jin, Xiaojie},
  journal={arXiv preprint arXiv:2406.08085},
  year={2024}
}

@article{li2024llava,
  title={Llava-onevision: Easy visual task transfer},
  author={Li, Bo and Zhang, Yuanhan and Guo, Dong and Zhang, Renrui and Li, Feng and Zhang, Hao and Zhang, Kaichen and Zhang, Peiyuan and Li, Yanwei and Liu, Ziwei and others},
  journal={arXiv preprint arXiv:2408.03326},
  year={2024}
}

@article{shen2024longvu,
  title={Longvu: Spatiotemporal adaptive compression for long video-language understanding},
  author={Shen, Xiaoqian and Xiong, Yunyang and Zhao, Changsheng and Wu, Lemeng and Chen, Jun and Zhu, Chenchen and Liu, Zechun and Xiao, Fanyi and Varadarajan, Balakrishnan and Bordes, Florian and others},
  journal={arXiv preprint arXiv:2410.17434},
  year={2024}
}

@inproceedings{qian2025dispider,
  title={Dispider: Enabling video llms with active real-time interaction via disentangled perception, decision, and reaction},
  author={Qian, Rui and Ding, Shuangrui and Dong, Xiaoyi and Zhang, Pan and Zang, Yuhang and Cao, Yuhang and Lin, Dahua and Wang, Jiaqi},
  booktitle={Proceedings of the Computer Vision and Pattern Recognition Conference},
  pages={24045--24055},
  year={2025}
}

@inproceedings{yao2025timechat,
  title={Timechat-online: 80\% visual tokens are naturally redundant in streaming videos},
  author={Yao, Linli and Li, Yicheng and Wei, Yuancheng and Li, Lei and Ren, Shuhuai and Liu, Yuanxin and Ouyang, Kun and Wang, Lean and Li, Shicheng and Li, Sida and others},
  booktitle={Proceedings of the 33rd ACM International Conference on Multimedia},
  pages={10807--10816},
  year={2025}
}

@article{zeng2026streamforest,
  title={Streamforest: Efficient online video understanding with persistent event memory},
  author={Zeng, Xiangyu and Qiu, Kefan and Zhang, Qingyu and Li, Xinhao and Wang, Jing and Li, Jiaxin and Yan, Ziang and Tian, Kun and Tian, Meng and Zhao, Xinhai and others},
  journal={Advances in Neural Information Processing Systems},
  volume={38},
  pages={75804--75835},
  year={2026}
}

@inproceedings{zhang2026hermes,
  title={Hermes: Kv cache as hierarchical memory for efficient streaming video understanding},
  author={Zhang, Haowei and Yang, Shudong and Fu, Jinlan and Ng, See Kiong and Qiu, Xipeng},
  booktitle={Proceedings of the 64th Annual Meeting of the Association for Computational Linguistics (Volume 1: Long Papers)},
  pages={8411--8430},
  year={2026}
}

@article{shen2026simple,
  title={A simple baseline for streaming video understanding},
  author={Shen, Yujiao and Tian, Shulin and Yang, Jingkang and Liu, Ziwei},
  journal={arXiv preprint arXiv:2604.02317},
  year={2026}
}

@article{lambert2024tulu,
  title={Tulu 3: Pushing frontiers in open language model post-training},
  author={Lambert, Nathan and Morrison, Jacob and Pyatkin, Valentina and Huang, Shengyi and Ivison, Hamish and Brahman, Faeze and Miranda, Lester James V and Liu, Alisa and Dziri, Nouha and Lyu, Shane and others},
  journal={arXiv preprint arXiv:2411.15124},
  year={2024}
}

@article{liu2026thinking,
  title={Thinking in streaming video},
  author={Liu, Zikang and Guo, Longteng and Li, Handong and Zhen, Ru and He, Xingjian and Ji, Ruyi and Ren, Xiaoming and Zhang, Yanhao and Lu, Haonan and Liu, Jing},
  journal={arXiv preprint arXiv:2603.12938},
  year={2026}
}

@article{yang2025longvt,
  title={LongVT: Incentivizing ``Thinking with Long Videos'' via Native Tool Calling},
  author={Yang, Zuhao and Wang, Sudong and Zhang, Kaichen and Wu, Keming and Leng, Sicong and Zhang, Yifan and Li, Bo and Qin, Chengwei and Lu, Shijian and Li, Xingxuan and others},
  journal={arXiv preprint arXiv:2511.20785},
  year={2025}
}

@article{yang2026paravt,
  title={ParaVT: Taming the Tool Prior Paradox for Parallel Tool Use in Agentic Video Reinforcement Learning},
  author={Yang, Zuhao and Zhang, Kaichen and Wang, Sudong and Wu, Keming and Yang, Zhongyu and Li, Bo and Qi, Xiaojuan and Lu, Shijian and Li, Xingxuan and Bing, Lidong},
  journal={arXiv preprint arXiv:2605.20342},
  year={2026}
}

@article{zhang2025openmmreasoner,
  title={OpenMMReasoner: Pushing the Frontiers for Multimodal Reasoning with an Open and General Recipe},
  author={Zhang, Kaichen and Wu, Keming and Yang, Zuhao and Li, Bo and Hu, Kairui and Wang, Bin and Liu, Ziwei and Li, Xingxuan and Bing, Lidong},
  journal={arXiv preprint arXiv:2511.16334},
  year={2025}
}

@article{zhao2026self,
  title={Self-Distilled Reasoner: On-Policy Self-Distillation for Large Language Models},
  author={Zhao, Siyan and Xie, Zhihui and Liu, Mengchen and Huang, Jing and Pang, Guan and Chen, Feiyu and Grover, Aditya},
  journal={arXiv preprint arXiv:2601.18734},
  year={2026}
}

@article{tian2026vicur,
  title={ViCuR: Visual Cues as Recoverable Privilege for Multimodal On-Policy Distillation},
  author={Tian, Kanghui and Liu, Siyuan and Yan, Ziang and Xia, Sheng and Dong, Shuai and Wang, Yi},
  journal={arXiv preprint arXiv:2606.05718},
  year={2026}
}

@article{sun2026v,
  title={V-Zero: Answer-Label-Free On-Policy Distillation with Contrastive Evidence Gating for Fine-Grained Visual Reasoning},
  author={Sun, Haoxiang and Yi, Zhihang and Deng, Langxuan and Zhou, Yuhao and Jia, Peiqi and Zhao, Jian and Yuan, Li and Lv, Jiancheng and Wang, Tao},
  journal={arXiv preprint arXiv:2606.25319},
  year={2026}
}

@article{li2026video,
  title={Video-OPD: Efficient Post-Training of Multimodal Large Language Models for Temporal Video Grounding via On-Policy Distillation},
  author={Li, Jiaze and Yin, Hao and Xu, Haoran and Xu, Boshen and Tan, Wenhui and He, Zewen and Ju, Jianzhong and Luo, Zhenbo and Luan, Jian},
  journal={arXiv preprint arXiv:2602.02994},
  year={2026}
}

@article{wang2026beyond,
  title={Beyond SFT-to-RL: Pre-alignment via Black-Box On-Policy Distillation for Multimodal RL},
  author={Wang, Sudong and Huang, Weiquan and Yu, Xiaomin and Yang, Zuhao and Lin, Hehai and Wu, Keming and Xiao, Chaojun and Chen, Chen and Wang, Wenxuan and Zhu, Beier and others},
  journal={arXiv preprint arXiv:2604.28123},
  year={2026}
}

@inproceedings{streamingbench,
  title={StreamingBench: Assessing the Gap for MLLMs to Achieve Streaming Video Understanding},
  author={Lin, Junming and Fang, Zheng and Chen, Chi and Wan, Zihao and Luo, Fuwen and Li, Peng and Liu, Yang and Sun, Maosong},
  year={2024},
  eprint={2411.03628},
  archivePrefix={arXiv},
  primaryClass={cs.CV},
}

@inproceedings{longvideobench,
  title={LongVideoBench: A Benchmark for Long-context Interleaved Video-Language Understanding},
  author={Wu, Haoning and Li, Dongxu and Chen, Bei and Li, Junnan},
  booktitle={Advances in Neural Information Processing Systems (NeurIPS)},
  year={2024},
}

\clearpage

\appendix

\section{Implementation and Data Details}

\paragraph{Systems Implementation.}
We build on a HybridFlow-style RL framework~\citep{sheng2025hybridflow} with vLLM~\citep{kwon2023efficient} rollouts, and run all training experiments on NVIDIA H200 GPUs. Four components were customized for video OPD and teacher privilege: (1) response-only teacher log-probability extraction, which aligns teacher and student despite differing prompt lengths; (2) a corrupt-video-tolerant data path that skips unreadable clips; (3) verifiable reward functions for MCQ, binary, and counting formats; and (4) an independent teacher-input builder that supports different frame budgets or prompts. For ST-CueGate, the builder issues the second no-cue teacher forward on the same teacher pool while reusing the student's decoded frames. Samples are routed to teachers by data source, and all privilege paths default to a no-op when disabled.

\paragraph{Training Configuration.}
Table~\ref{tab:training-config} lists the configuration used by the main ST-CueGate run. Standard OPD shares the same optimizer, sequence limits, and model pair, but uses $n{=}1$ and disables teacher-side cue conditioning and gating.

\begin{table}[t]
\centering
\small
\setlength{\tabcolsep}{4pt}
\begin{tabular}{@{}lp{0.57\columnwidth}@{}}
\toprule
\textbf{Parameter} & \textbf{Value} \\
\midrule
Student / teacher & Qwen3.5-4B / Qwen3.5-9B \\
Training / evaluation mode & thinking / instruct \\
Training visual input & default 2\,fps; no explicit frame cap \\
Training data & 25,118 cueinstruct samples \\
Epochs & 4 \\
Random seeds & 42, 43, 44 \\
Batch / mini-batch size & 32 / 32 \\
Learning rate & $1\times10^{-6}$ \\
Rollouts per prompt & $n=4$ \\
Max prompt / response length & 16,000 / 512 tokens \\
Distillation & sampled-token k1 + clipped policy gradient; top-$k=64$ \\
Gate & sequence level; UID grouping; $\alpha_{\mathrm{g}}=0.5$; $\gamma=1$; $w\in[0,2]$ \\
GPU allocation & 4 student + 4 teacher NVIDIA H200 GPUs \\
Checkpoint frequency & every 100 steps \\
\bottomrule
\end{tabular}
\caption{Training configuration for the main ST-CueGate experiment.}
\label{tab:training-config}
\end{table}

\paragraph{Searched Values and Selection Criterion.}
Table~\ref{tab:search-space} lists every value we tried per axis together with the setting used in the main results. Axes were explored one at a time from the default configuration rather than through a joint grid search. Selection always uses the held-out validation aggregate; no downstream benchmark score was consulted when choosing a setting.

\begin{table}[t]
\centering
\small
\setlength{\tabcolsep}{4pt}
\begin{tabular}{@{}lp{0.40\columnwidth}l@{}}
\toprule
\textbf{Axis} & \textbf{Values tried} & \textbf{Used} \\
\midrule
Gate strength $\alpha_{\mathrm{g}}$ & $0.25$, $0.5$, $1.0$ & $0.5$ \\
Rollouts per prompt $n$ & $1$, $4$, $8$ & $4$ \\
Gate granularity & sequence, token & sequence \\
Comparison group & minibatch, prompt UID & prompt UID \\
Teacher & Qwen3.5-9B, Qwen3.5-27B, frozen 4B copy & Qwen3.5-9B \\
Learning rate & $1\times10^{-6}$, $2\times10^{-7}$ & $1\times10^{-6}$ \\
Training pool & 8.3k, 20k, 25k, 26k, 57k & 25k \\
ExOPD $\lambda$ & $1.0$, $1.25$; two-axis $\lambda_t{=}1.5,\lambda_c{=}1.0$ & not used \\
DAD $\alpha_{\mathrm{DAD}}$ & $1.0$ & not used \\
\bottomrule
\end{tabular}
\caption{Values explored per hyperparameter and the setting used in the reported runs. Gate range $[w_{\min},w_{\max}]{=}[0,2]$, $\gamma{=}1$, and four epochs were fixed throughout.}
\label{tab:search-space}
\end{table}

\paragraph{Same-Backbone OPSD Instantiation.}
For ST-CueGate--OPSD, we replace the 9B teacher with a frozen copy of the 4B student's initial policy, following the same-model teacher--student construction of OPSD~\citep{zhao2026self}. The frozen copy receives the cue and no-cue teacher contexts, while the updating student retains its original prompt; the remaining training and evaluation settings are unchanged.

\paragraph{Evaluation Protocol.}
Table~\ref{tab:evaluation-protocol} summarizes the per-benchmark settings. All checkpoints are evaluated with instruct-mode greedy decoding. StreamingBench and OVO-Bench use the deployment-aligned recent-window protocol, whereas the two general-video benchmarks follow their standard evaluation settings. For OVO-Bench, we report the full macro, the score excluding HLD, and raw HLD. The separation follows SimpleStream's observation that HLD tests hallucination robustness rather than episodic event recall~\citep{shen2026simple}; it does not remove HLD from the deployment analysis.

\begin{table*}[t]
\centering
\small
\setlength{\tabcolsep}{5pt}
\begin{tabular}{@{}llll@{}}
\toprule
\textbf{Benchmark} & \textbf{Visual Input} & \textbf{Decoding} & \textbf{Reported Metric} \\
\midrule
StreamingBench & recent 4 frames at 1\,fps & instruct, greedy & answer accuracy \\
OVO-Bench & recent 4 frames at 1\,fps & instruct, greedy & backward+realtime macro, with/without HLD \\
Video-MME & standard lmms-eval, max 32 frames & instruct, greedy & overall accuracy \\
LongVideoBench & standard lmms-eval, max 32 frames & instruct, greedy & overall accuracy \\
\bottomrule
\end{tabular}
\caption{Evaluation protocols. The recent-window constraint applies to the streaming benchmarks; Video-MME and LongVideoBench use their standard general-video protocols.}
\label{tab:evaluation-protocol}
\end{table*}

Model selection uses only a held-out validation aggregate, never downstream test scores. Each run contributes one selected model, and that same model is evaluated on all four benchmarks and all reported subtasks. Distributed checkpoints are merged to Hugging Face format before evaluation. The repository entry points are \texttt{scripts/eval\_all.sh} for the four benchmark jobs and \texttt{scripts/score\_all.sh} for aggregation.

\paragraph{Statistical Reporting.}
Each in-house trained configuration uses seeds 42, 43, and 44. Each run contributes a single model selected on the held-out validation aggregate, which is then evaluated on all four benchmarks, and table entries are arithmetic means over the three selected models. Only aggregate means were retained, so standard deviations and paired significance tests are unavailable. Deterministic greedy decoding removes sampling noise at evaluation but not training or checkpoint variance.

\section{Recent-Window vs.\ Dense-Uniform Context}

We additionally evaluate a single fixed ST-CueGate model under two causal context policies. The main protocol observes the most recent four frames at 1\,fps. The dense-uniform protocol instead samples up to 32 frames uniformly from the full visible prefix $[0,t]$, matching the frame-count convention used by standard offline video evaluation. All other decoding and scoring settings are unchanged.

\begin{table}[t]
\centering
\small
\setlength{\tabcolsep}{3pt}
\begin{tabular}{@{}lrrr@{}}
\toprule
\textbf{Metric} & \textbf{Recent-4} & \textbf{Dense-32} & \textbf{$\Delta$} \\
\midrule
StreamingBench             & 84.19 & 81.15 & $-3.04$ \\
OVO incl.\ HLD             & 68.29 & 63.06 & $-5.23$ \\
OVO excl.\ HLD             & 70.48 & 69.03 & $-1.45$ \\
Backward Avg.\ incl.\ HLD  & 54.47 & 54.51 & $+0.04$ \\
Backward Avg.\ excl.\ HLD  & 58.85 & \textbf{66.45} & $\mathbf{+7.60}$ \\
Realtime Avg.              & 82.11 & 71.61 & $-10.50$ \\
EPM                        & 54.47 & \textbf{61.95} & $\mathbf{+7.48}$ \\
ASI                        & 58.78 & \textbf{70.95} & $\mathbf{+12.17}$ \\
HLD                        & 45.70 & 30.65 & $-15.05$ \\
\bottomrule
\end{tabular}
\caption{Context-policy comparison on the same fixed ST-CueGate model. Dense-32 uniformly samples the visible prefix, while Recent-4 retains only the latest four frames. EPM: Episodic Memory; ASI: Action Sequence Identification; HLD: Hallucination Detection.}
\label{tab:context-policy}
\end{table}

Table~\ref{tab:context-policy} reveals a clear perception--memory trade-off. Dense historical context substantially improves backward memory reasoning, especially ASI ($+12.17$) and EPM ($+7.48$), but dilutes recent-event evidence: the realtime macro drops by $10.50$ points and HLD by $15.05$ points. StreamingBench likewise decreases by $3.04$ points. Thus, dense-uniform context is useful for historical recall but is not uniformly better for streaming; the recent-window protocol deliberately prioritizes current-event perception and deployment efficiency.

\section{Prompt Templates}

This section reproduces verbatim every prompt used for training, cue construction, and evaluation. Curly braces mark runtime substitutions.

\paragraph{Student Prompt.}
The student prompt is identical during training and deployment and never contains a cue.
\begin{promptbox}{Student prompt (training and deployment)}
\begin{lstlisting}
<video>{question}
A. {option A}
B. {option B}
C. {option C}
D. {option D}

(Binary and counting items carry the question only, with no option block.)
\end{lstlisting}
\end{promptbox}

\paragraph{Cue-Augmented Teacher Prompt.}
For cue-only ViCuR and ST-CueGate, the teacher receives the unchanged student content plus an explicit instruction block. The framing follows reference-solution prompting~\citep{zhao2026self}, with the reference solution replaced by visual evidence.
\begin{promptbox}{Teacher prompt with spatio-temporal cue}
\begin{lstlisting}
<video>{question}
A. {option A}
B. {option B}
C. {option C}
D. {option D}

Here is some visual evidence to help you locate the answer in the video:
{cue}

After considering this visual evidence, identify the exact moment or region it points to in the video, verify what actually happens there, then answer the question using your own judgment.
\end{lstlisting}
\end{promptbox}
The ST-CueGate no-cue forward reuses this same teacher, the decoded student frames, and the original student prompt. Fallback samples without an accepted cue therefore satisfy $\tau^+=\tau^-$, $\Delta=0$, and a neutral gate.

\paragraph{Cue Generator.}
The generator sees the complete training clip and the question stem after answer options are stripped from its input. Multiple-choice items use the following system prompt.
\begin{promptbox}[colback=pbGreen, colframe=pbGreenEdge, colbacktitle=pbGreenEdge]{Cue generator: multiple-choice items}
\begin{lstlisting}
You are a video analyst. For a multiple-choice question about a video, output a POINTER that tells the student WHEN and WHERE to look -- WITHOUT answering the question.
Think of it as pointing a flashlight: you reveal the location and moment, never the finding.

ABSOLUTE PROHIBITIONS (any violation = total failure):
1. NEVER output an option letter or option text (no 'A)', 'B.', 'E) To ...', and no sentence that restates any option).
2. NEVER name the specific entity/attribute the question asks about. If it asks WHO/WHICH PERSON, don't name the person; if WHERE/POSITION, don't state the position; if WHAT OBJECT, don't name the object; if WHAT COLOR, don't say the color; if WHAT HAPPENS NEXT, don't say the outcome. Refer only generically ('that person', 'the object in their hand', 'their position').
3. NEVER quote or transcribe on-screen text, captions, subtitles, or spoken dialogue. Only say WHEN/WHERE text appears (e.g. 'a subtitle appears around 0:09'), never its wording.
4. NEVER describe the outcome, result, or a conclusion. Pointer only, no interpretation.

FORMAT: exactly ONE sentence, under 25 words, shape = 'Around <when/which action>, look at <which region/subject>.'  No extra clauses, no 'suggesting', no 'indicating'.

GOOD (Q: what color are the sneakers when the hand enters?):
  'Around the moment the hand first enters the frame, look at the person's feet.'
GOOD (Q: who places a tile next after the man with glasses?):
  'Right after the man with glasses moves, look at the other players seated around the board.'
GOOD (Q: where is the red bowl relative to the pizza?):
  'As the woman slices the pizza, look at the area surrounding the cutting board.'
BAD -- all leak, never do these:
  'E) To make sure the image is captured.'  (option letter+text)
  'The girl in the purple shirt.'  (names the answer entity)
  'To the right of the pizza.'  (states the position asked)
  'The caption says "Pregnant".'  (transcribes text)
  '...standing together, suggesting they are on a break.'  (describes outcome/conclusion)
\end{lstlisting}
\end{promptbox}
Binary items embed the queried attribute in the question itself, so any description of the evidence risks touching the answer. They therefore use a stricter variant that degenerates the cue into a pure spatio-temporal pointer.
\begin{promptbox}[colback=pbGreen, colframe=pbGreenEdge, colbacktitle=pbGreenEdge]{Cue generator: binary items}
\begin{lstlisting}
You are a video analyst helping a student LOCATE where to look for a yes/no question, WITHOUT answering it.
The question asks whether some attribute/event holds at a moment. Your job is ONLY to point to WHEN and WHO/WHERE to inspect -- NEVER to say whether it holds.

STRICT RULES (violating any = failure):
1. NEVER say yes/no, true/false, is/is not, was/wasn't, there is/there isn't, correct/incorrect, or any confirmation or denial.
2. NEVER mention the specific attribute the question asks about (e.g. if it asks 'blue helmet', do NOT mention any helmet color; if 'red grinder', do NOT mention any grinder color). Refer to the subject only generically (the helmet, the tool, the person's hand).
3. NEVER describe the state/outcome. Only give a spatio-temporal pointer.
4. Output ONE sentence, under 25 words, of the form: 'Around <when/which action>, look at <who/which region>.'

Good example (Q: Is the child in the red canoe wearing a blue helmet?): "Around the segment showing the child in the red canoe, look closely at the head area."
Bad examples (all leak): "No, the helmet is yellow." / "The child wears a red helmet." / "Yes, there is a blue helmet."
\end{lstlisting}
\end{promptbox}

\paragraph{Regeneration and Automatic Judging.}
When rule-based screening flags a candidate, the generator is re-queried with the original messages, its flagged output, and the instruction below, at temperatures $0.5$, $0.7$, and $0.9$.
\begin{promptbox}[colback=pbPurple, colframe=pbPurpleEdge, colbacktitle=pbPurpleEdge]{Cue regeneration after a flagged candidate}
\begin{lstlisting}
Your previous response LEAKED the answer (it stated or restated an option/answer/color/count/yes-no/on-screen text). Rewrite it as a PURE POINTER: one short sentence 'Around <when/action>, look at <which region/subject>.' Do NOT state or hint the answer, do NOT output any option letter, and do NOT quote on-screen text. Output only the pointer.
\end{lstlisting}
\end{promptbox}
A temperature-zero pass of the same model then judges the question--answer--cue tuple and emits a single word.
\begin{promptbox}[colback=pbPurple, colframe=pbPurpleEdge, colbacktitle=pbPurpleEdge]{Automatic leakage judge}
\begin{lstlisting}
You grade whether a HINT leaks the answer to a video question. A good hint ONLY points to WHERE/WHEN to look; it may name a region, object, or moment, but must not reveal the ANSWER.

Judge by this rule:
- SAFE: the hint is a pure pointer of the form 'look at <region/subject> around <when>'. It is still SAFE even if it names the object/person/area being asked about, AS LONG AS it does NOT state the answered attribute (the color, the count, the yes/no, the identity, the outcome, or the on-screen text content).
- LEAK: the hint states or strongly implies the answer -- e.g. gives the color/count/yes-no, names WHICH specific person/object is the answer, quotes the on-screen text that IS the answer, or describes the outcome/next-action the question asks for.

Examples:
  Q: where is the chair relative to the basket? Hint: 'look at the object directly underneath the basket.' -> SAFE (points to a location to inspect, does not state the spatial answer)
  Q: what does the on-screen text say? Hint: "the text 'Pregnant' is visible" -> LEAK (quotes the answer text)
  Q: what happens next? Hint: 'she rolls forward and rises, preparing for the next drill' -> LEAK (describes the outcome asked)
  Q: next action of the bull? Hint: 'look at the bull's body orientation and the man's posture' -> SAFE (points to what to observe, does not state the action)

Reply with EXACTLY one word: LEAK or SAFE.

--- user message ---
Question: {question}
Correct answer: {ground truth}
Hint: {cue}

Verdict (LEAK or SAFE):
\end{lstlisting}
\end{promptbox}

\paragraph{Evaluation Prompts.}
StreamingBench items are presented with an explicit direct-answer instruction.
\begin{promptbox}[colback=pbAmber, colframe=pbAmberEdge, colbacktitle=pbAmberEdge]{Evaluation: StreamingBench}
\begin{lstlisting}
You are an advanced video question-answering AI assistant. You have been provided with some frames from the video and a multiple-choice question. Your task is to analyze the video and provide the best answer.

Question: {question}

Options:
{options}

Only give the best option's letter (A, B, C, or D) directly.
\end{lstlisting}
\end{promptbox}
The OVO-Bench backward-tracing and real-time subtasks use a shorter template with the same direct-answer constraint.
\begin{promptbox}[colback=pbAmber, colframe=pbAmberEdge, colbacktitle=pbAmberEdge]{Evaluation: OVO-Bench backward and real-time subtasks}
\begin{lstlisting}
{question}
Options: {options}
Only give the best option's letter directly.
\end{lstlisting}
\end{promptbox}
Video-MME and LongVideoBench keep the unmodified lmms-eval prompts of their standard protocols.

\section{Cue Construction and Screening}

\paragraph{Offline Cue Generation.}
The cue pipeline accepts any sufficiently strong frozen video-language model as its generator; we use Qwen3.5-27B in our implementation. The generator processes the complete short training clip and produces one spatio-temporal pointer per sample. It is instructed to output a single sentence of at most 25 words in the form ``Around \{when/action\}, look at \{region/subject\},'' without option letters, answer entities or attributes, quoted on-screen text, binary labels, or outcome descriptions. Thus, the cue is grounded at the level of the underlying training clip; it is not generated from the recent-4-frame evaluation window and is never provided at deployment.

\paragraph{Leakage Taxonomy, Automatic Screening, and Manual Audit.}
We operationally treat a cue as leaking if it reveals the answer without requiring the intended visual inspection. The audit taxonomy covers (1) exact answer strings or option identifiers, (2) synonyms and semantic paraphrases, (3) discriminative attributes such as color, count, identity, or position, (4) event or outcome paraphrases, and (5) clues that eliminate alternatives and make the correct option effectively identifiable. Rule-based checks cover option letters, binary assertions, explicit counts, quoted answer text, and lexical overlap with the correct MCQ option (word Jaccard at least $0.35$, or at least two answer keywords covering $60\%$ of its content words). Candidates that fail are regenerated up to three times with an explicit answer-withholding reminder and temperatures $0.5$, $0.7$, and $0.9$. A temperature-zero pass of the same generator model then classifies the question--ground-truth--cue tuple as \texttt{SAFE} or \texttt{LEAK}; judge failures are conservatively rejected.

Of 25,118 samples, 24,257 ($96.6\%$) are automatically accepted and 861 ($3.4\%$) fall back to the original teacher prompt. Because generation and automatic judging use the same model, we complement this self-check with a random manual audit of 300 accepted cues. Two annotators jointly inspect the video, question, options, stored answer, and cue for the five leakage categories above as well as whether the pointer is visually grounded. The audit flags $3$ direct-answer leaks (category 1), $4$ indirect semantic leaks (categories 2--5), and $6$ visually unsupported or incorrectly grounded pointers, i.e.\ $13$ of the $300$ sampled cues. Both annotators review each sample together and resolve disagreements by discussion, so these are consensus counts and we do not report an inter-annotator agreement coefficient.

Treating leakage as the union of the first two groups gives a residual leakage rate of $7/300=2.3\%$ among accepted cues ($95\%$ Wilson interval $1.1$--$4.7\%$), so the accept decision has an estimated precision of ${\sim}97.7\%$ and the end-to-end non-leaking fraction of the training pool is approximately $94\%$. Because the audit samples only accepted cues, it estimates this precision but not the false-rejection rate on the $861$ fallback samples. The audit is a post-hoc quality estimate and does not retroactively re-filter the training pool; $96.6\%$ therefore remains an automatic acceptance rate rather than a human-verified clean rate.

\paragraph{Representative Accepted Cues.}
Table~\ref{tab:cue-examples} lists cues sampled from the released training pool, two per answer format. They illustrate the intended behaviour of the screening rules. The binary example about strap colour points at the wrists without naming a colour; the counting example points at a region without stating a count; and the multiple-choice examples localise a moment and a subject without restating any option. Each cue is short and purely deictic, so the teacher still has to inspect the clip to answer.

\begin{table*}[t]
\centering
\small
\begin{tabular}{@{}llp{0.36\textwidth}p{0.38\textwidth}@{}}
\toprule
\textbf{Format} & \textbf{Answer} & \textbf{Question} & \textbf{Accepted cue} \\
\midrule
Multiple choice & C & As the camera focuses on the man in the tank top, what specific exercise is he performing? & Around 0:04, when the man in the tank top is shown, look at the movement of his arms and the cable machine he is using. \\
Multiple choice & D & As the man finishes sniffing the collar of his striped shirt, what is he about to do next? & Around 0:09, as he lowers his hand from his collar, look at his face and the direction he turns his head. \\
Binary & Yes & As the man wraps his wrists before lifting, are the straps he is using black and purple? & Around the segment where the man is preparing to lift, look closely at the straps on his wrists. \\
Binary & Yes & As the person is shown laughing, are their eyes closed? & Around the segment from 6.5 to 7.5 seconds, look at the person's face. \\
Counting & 0 & As the woman in the blue dress is laughing, how many lollipops are visible? & Around the time the woman in the blue dress is laughing, look at the hand in the foreground. \\
Counting & 2 & Just before the scene transitions, how many men are giving a thumbs-up gesture? & Around the moment the second text overlay appears, look at the hands of both men standing in the frame. \\
\bottomrule
\end{tabular}
\caption{Representative cues drawn from the accepted training pool, two per answer format. The answer column is shown here for inspection only; it is never part of the teacher prompt.}
\label{tab:cue-examples}
\end{table*}

\section{Answer Parsing and Verifiable Reward}

Ground truths are typed automatically: a single letter A--H marks a multiple-choice item, \texttt{yes}/\texttt{no} marks a binary item, and a non-negative integer marks a counting item. Each type has its own extractor, and the reward is $1$ on an exact match after extraction and $0$ otherwise. This rule-based reward drives only the GRPO diagnostic; the OPD objective never consults it.

Multiple-choice extraction first looks for an explicit answer cue such as ``the answer is'' and takes the last such occurrence, since models frequently restate. Absent a cue, only upper-case standalone letters are accepted, which prevents the English article ``a'' and the pronoun ``I'' from being read as options, and the \emph{last} qualifying letter wins. Binary extraction takes the \emph{first} \texttt{yes}/\texttt{no} token, because binary answers normally lead the response. Counting extraction prefers the first integer after an answer cue and otherwise the first integer in the response.

The direction of the multiple-choice rule matters for the format-drift diagnostic. The training reward reads the last option letter, whereas a deployed streaming parser reads the first answer token it encounters. A model that emits a long rationale before its answer scores well under the former and poorly under the latter, which is precisely the gap reported for GRPO in the main text. Distilled models emit a bare option letter, so both parsers agree.

\section{Training Algorithm}

Algorithm~\ref{alg:st-cuegate} gives the per-batch procedure for ST-CueGate training, including the two teacher replays and the group-relative gate.

\begin{algorithm}[t]
\caption{ST-CueGate Training}
\label{alg:st-cuegate}
\begin{algorithmic}[1]
\REQUIRE batch of prompts $q_i$, cues $c_i$, student $\pi_\theta$, frozen teacher $\pi_\tau$
\FOR{each prompt $q_i$}
    \STATE Sample an on-policy response $y_i\sim\pi_\theta(\cdot\mid q_i)$ in thinking mode.
    \STATE Compute student token log-probabilities $s_{i,t}$ on response tokens.
    \STATE Replay $y_i$ with the teacher under the cue prompt to obtain $\tau^+_{i,t}$.
    \STATE Replay $y_i$ with the same teacher and frames under the original prompt to obtain $\tau^-_{i,t}$.
    \STATE Compute $\Delta_{i,t}=\tau^+_{i,t}-\tau^-_{i,t}$ and $g_i=T_i^{-1}\sum_t\Delta_{i,t}$.
\ENDFOR
\STATE Standardize $g_i$ within each prompt's sibling rollout group.
\STATE Set $w_i=\mathrm{clip}(1+\alpha_{\mathrm{g}}\widetilde g_i,w_{\min},w_{\max})$.
\STATE Form $A_{i,t}^{\mathrm{ST\text{-}CueGate}}=w_i(\tau^+_{i,t}-s_{i,t})$.
\STATE Update $\pi_\theta$ with the clipped policy-gradient objective.
\end{algorithmic}
\end{algorithm}

\paragraph{Gate Interpretation and Unexplored Aggregations.}
The gate is a group-relative response reweight, not an absolute test of whether a cue helped. In particular, an above-mean response can receive $w_i>1$ even when all $g_i$ values in its sibling group are negative. The response mean also mixes reasoning, answer, punctuation, and formatting tokens, and the resulting scalar multiplies every token-level OPD advantage. We did not evaluate absolute positive-clipped gates, positive-token-only or answer-token-only aggregation, top-$k$/quantile/trimmed means, or hard suppression of negative-score responses. These alternatives could better separate absolute cue benefit from relative ranking and reduce sensitivity to response composition. Our results therefore establish the performance of one likelihood-based response proxy, not the faithfulness or optimality of its aggregation rule. The implementation additionally supports asymmetric penalization of negative-score tokens through a coefficient $\gamma$, but all reported runs use $\gamma=1$, which reduces that extension exactly to the length-normalized response mean defined in the main paper.

\section{Objective Derivations and Policy-Gradient View}

\paragraph{Notation.}
Let $q_i$ denote the $i$-th prompt, consisting of the decoded video frames and the question text, $\pi_\theta$ the student, $\pi_\tau$ the frozen teacher, and $\pi_b$ the frozen student initialization used by the extrapolation baselines. For a sampled response $y_i$ of length $T_i$ we write $s_{i,t}$, $\tau_{i,t}$, and $b_{i,t}$ for the corresponding token log-probabilities, as in the main text.

\paragraph{From Reverse KL to the Sampled-Token Estimator.}
On-policy distillation minimizes the reverse KL between student and teacher at every position along the student's own trajectory,
\begin{equation}
\mathrm{KL}_{i,t}=\!\!\sum_{v\in\mathcal{V}}\pi_\theta(v\mid y_{i,<t},q_i)\,
\log\frac{\pi_\theta(v\mid y_{i,<t},q_i)}{\pi_\tau(v\mid y_{i,<t},q_i)}.
\end{equation}
Summing over the full vocabulary at every position is expensive for long multimodal prompts, so we use the single-sample estimator evaluated at the realized token,
\begin{equation}
k^{(1)}_{i,t}=s_{i,t}-\tau_{i,t},
\end{equation}
which is unbiased for $\mathrm{KL}_{i,t}$ because $y_{i,t}\sim\pi_\theta(\cdot\mid y_{i,<t},q_i)$. The OPD advantage of the main text is exactly its negation, $A^{\mathrm{OPD}}_{i,t}=-k^{(1)}_{i,t}$. The implementation also offers the lower-variance $k^{(2)}$ and $k^{(3)}$ estimators, but all reported runs use $k^{(1)}$: it is signed, which is what makes the advantage reading below possible, whereas the low-variance estimators are non-negative by construction.

\paragraph{OPD as a Dense-Reward Policy Gradient.}
Because the advantage carries a stop-gradient, differentiating the token-level policy-gradient objective of the main text gives
\begin{equation}
\nabla_\theta\mathcal{L}_{\mathrm{PG}}
=-\mathbb{E}\!\left[\frac{1}{T_i}\sum_{t=1}^{T_i}
A_{i,t}\,\nabla_\theta\log\pi_\theta(y_{i,t}\mid y_{i,<t},q_i)\right],
\end{equation}
so OPD performs gradient ascent on the expected per-token reward $r_{i,t}=\tau_{i,t}-s_{i,t}$ along on-policy rollouts. This reward is dense and self-annihilating: it is positive exactly where the teacher places more mass than the student on the realized token, and it vanishes identically once the student matches the teacher on the sampled support. The GRPO diagnostic differs structurally on both counts. Its advantage is constant across all tokens of a response, and it vanishes whenever every rollout in a group receives the same verifiable reward, which is why sparse task reward can constrain the final answer without constraining the trajectory that precedes it.

\paragraph{Extrapolated Targets and Why They Are Ungrounded.}
ExOPD replaces the teacher log-probability with a point extrapolated along the teacher's improvement over the base, $b_{i,t}+\lambda(\tau_{i,t}-b_{i,t})$, which yields the ExOPD advantage given in the main text; $\lambda=1$ returns standard OPD. For $\lambda>1$ the extrapolated vector is in general not the log-density of any normalized distribution, since $\sum_{v}\exp[b+\lambda(\tau-b)]\neq1$, so no teacher policy realizes the target. The update therefore pushes the student toward a point that no achievable conditional distribution occupies, and, unlike a cue, the direction is not tied to anything observable in the training clip. This is the precise sense in which we call the signal ungrounded, and it is consistent with the instability reported below.

\paragraph{Axis-Decomposed Extrapolation.}
AD-ExOPD adds a second frozen reference $\pi_b^{\mathrm{dense}}$, the same base model given the teacher's dense frame budget, and splits the teacher's improvement into a frame axis and a capacity axis,
\begin{equation}
\tau-b=\underbrace{(b^{\mathrm{dense}}-b)}_{\text{more frames}}
+\underbrace{(\tau-b^{\mathrm{dense}})}_{\text{more capacity}},
\end{equation}
extrapolating each with its own gain:
\begin{equation}
A^{\mathrm{AD}}_{i,t}=-(s-b)+\lambda_{t}(b^{\mathrm{dense}}-b)+\lambda_{c}(\tau-b^{\mathrm{dense}}).
\end{equation}
Setting $\lambda_t=\lambda_c$ recovers ExOPD. We ran $\lambda_t{=}1.5$, $\lambda_c{=}1.0$, on the reasoning that a sliding-window student can exploit additional frames but cannot acquire 9B capacity; the outcome is reported below.

\paragraph{The Gate as a Length-Normalized Likelihood Ratio.}
Because both teacher passes score the same realized response, the token contrasts telescope into a sequence-level quantity:
\begin{equation}
\sum_{t=1}^{T_i}\Delta_{i,t}
=\log\pi_\tau(y_i\mid q_i,c_i)-\log\pi_\tau(y_i\mid q_i),
\end{equation}
so $g_i$ is the length-normalized log-likelihood ratio of one response under two nested teacher contexts. Algebraically this has the form of a pointwise mutual information between the cue and the realized response. We deliberately avoid that reading: the two conditionals are obtained by prompting a single network in two ways and need not be consistent with any joint distribution over cues and responses, so we treat $g_i$ as a likelihood-ratio statistic rather than an information-theoretic quantity.

\paragraph{The Gate Is Mean-Preserving, and When It Saturates.}
Standardizing within a group of $n$ sibling rollouts with population statistics gives $\sum_{j\in\mathcal{G}}\widetilde g_j=0$, so before clipping
\begin{equation}
\frac{1}{|\mathcal{G}|}\sum_{i\in\mathcal{G}}w_i
=1+\frac{\alpha_{\mathrm{g}}}{|\mathcal{G}|}\sum_{i\in\mathcal{G}}\widetilde g_i=1
\end{equation}
exactly. The gate therefore redistributes weight inside a group rather than rescaling the group's total contribution, and $\alpha_{\mathrm{g}}$ controls only the spread. Standardized scores are also bounded: for $n$ values, $|\widetilde g|\le\sqrt{n-1}$, with equality when one member differs from the other $n-1$. The attainable weight range is thus $[1-\alpha_{\mathrm{g}}\sqrt{n-1},\,1+\alpha_{\mathrm{g}}\sqrt{n-1}]$, summarized in Table~\ref{tab:gate-range}.

\begin{table}[t]
\centering
\small
\begin{tabular}{@{}llcc@{}}
\toprule
$n$ & $\alpha_{\mathrm{g}}$ & Attainable $w$ range & Clip reachable \\
\midrule
4 & 0.25 & $[0.57,\,1.43]$ & no \\
4 & 0.50 & $[0.13,\,1.87]$ & no \\
4 & 1.00 & $[-0.73,\,2.73]$ & yes \\
8 & 0.50 & $[-0.32,\,2.32]$ & yes \\
\bottomrule
\end{tabular}
\caption{Attainable gate values before clipping to $[w_{\min},w_{\max}]{=}[0,2]$, from the bound $|\widetilde g|\le\sqrt{n-1}$. The reported configuration ($n{=}4$, $\alpha_{\mathrm{g}}{=}0.5$) cannot reach either boundary.}
\label{tab:gate-range}
\end{table}

For the reported configuration the range lies strictly inside the clip interval, so the gate is provably a smooth monotone reweighting and the bounds never bind. Raising $\alpha_{\mathrm{g}}$ to $1.0$ at $n{=}4$, or widening the group to $n{=}8$ at $\alpha_{\mathrm{g}}{=}0.5$, makes both boundaries reachable, so a single extreme rollout can be suppressed to zero or doubled. Those are also the two settings that lose accuracy in the corresponding ablations. We report this as a consistency observation rather than an identified cause, since changing $n$ and $\alpha_{\mathrm{g}}$ also changes the variance of the standardized score itself.

\paragraph{A Common Reweighting Family.}
Standard OPD, the ungrounded controls, and the contrastive gates are all instances of
\begin{equation}
A_{i,t}=w_i\,(\tau_{i,t}-s_{i,t}),
\end{equation}
and differ only in what generates $w_i$: it is constant for OPD; derived from student--teacher disagreement $|s-\tau|$ for DAD, which uses no external context; derived from a positive and a degraded visual view for the V-Zero adaptation; and derived from cue and no-cue teacher contexts over the same frames and response for ST-CueGate. The reweighting mechanism is therefore shared, and our contribution lies in the choice of reference, not in the existence of a group-relative gate.

\section{Data Construction and Statistics}
\begin{figure*}[t]
\centering
\includegraphics[width=0.9\textwidth]{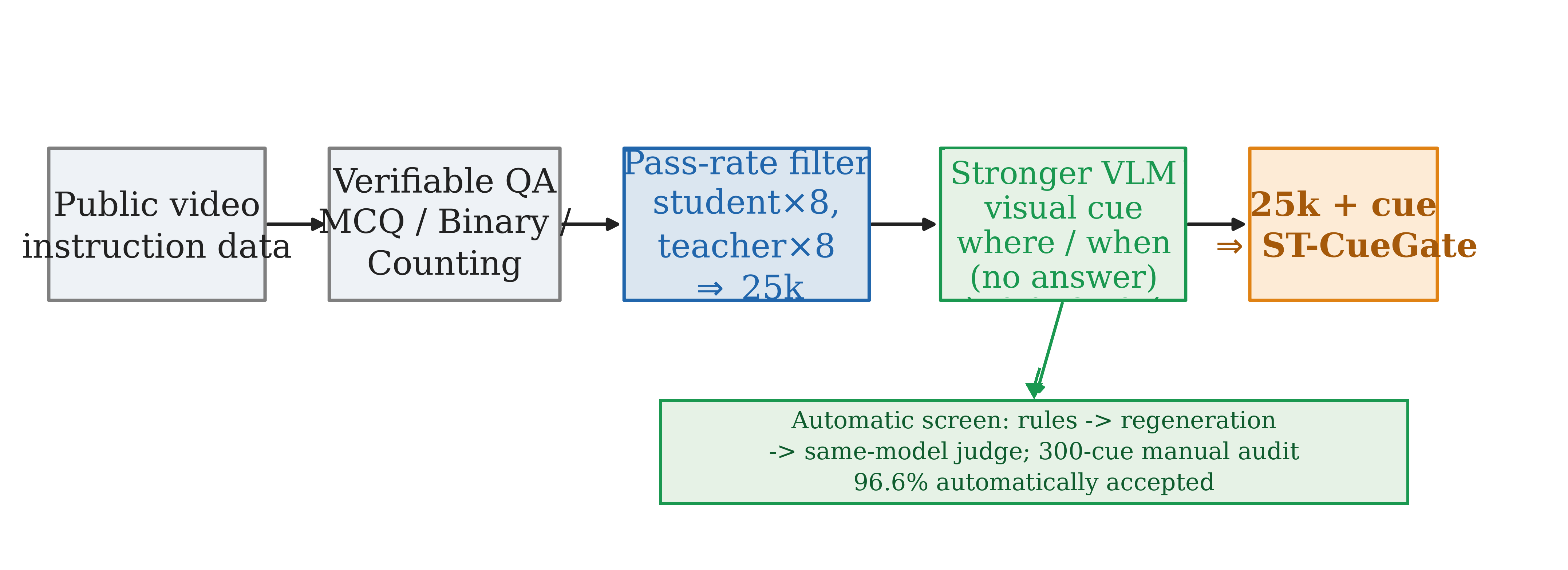}
\caption{Data construction pipeline. Public video instruction data are cast into verifiable QA, pass-rate filtered by $8$-sample student/teacher agreement, and annotated offline by a stronger VLM with an automatically screened spatio-temporal visual cue. The cue is wrapped in an explicit instruction frame and appended only to the teacher prompt.}
\label{fig:datapipe}
\end{figure*}

Training data are streaming-video QA items constructed from public instruction data~\citep{zhang2024llava,yuan2025tarsier2}. Videos are short (median ${\sim}5.5$s, $89\%$ under $15$s) and heavily temporal ($71.8\%$ of questions require temporal reasoning); $69.7\%$ are MCQ, $25.3\%$ binary, and $5.0\%$ counting. Figure~\ref{fig:datapipe} summarizes the pipeline and Table~\ref{tab:data} gives the exact data lineage.

\paragraph{Pass-Rate Filtering.}
We sample the student and teacher eight times each at temperature $0.7$ and estimate per-sample pass rates. Removing samples that are already easy for both models yields an $8{,}343$-item filtered set. Among the $4.6\%$ of samples where the teacher underperforms the student, $40.7\%$ are response-format deviations rather than genuine capability gaps.

\begin{table}[t]
\centering
\begin{tabular}{lrl}
\toprule
Dataset & Size & Composition \\
\midrule
8.3k filtered       & 8{,}343  & 20k after pass-rate filter \\
20k full            & 20{,}031 & rlvr $\cup$ cot\_rlvr \\
cot\_verifiable     & 23{,}884 & verifiable CoT pool \\
\textbf{25k merged} & 25{,}118 & 8.3k $\cup$ cot\_verifiable \\
Cue injected        & 24{,}257 & automatically accepted teacher cue \\
Cue fallback        & 861      & original teacher prompt \\
\bottomrule
\end{tabular}
\caption{Training-data construction. The 25k merged set is the default training configuration.}
\label{tab:data}
\end{table}

\section{Cue-Conditional Contrast Analysis}

\paragraph{Collection Protocol.}
For each sampled student response, we retain only valid response-token positions and compute $\Delta_t=\log p_T(y_t\mid\mathrm{cue})-\log p_T(y_t\mid\mathrm{no\text{-}cue})$ using the same teacher and decoded frames. The main distribution pools 300,866 tokens collected over five diagnostic training steps. We classify tokens with $|\Delta_t|\leq0.01$ as inactive and compute concentration after sorting positive-$\Delta$ tokens by score.

\begin{figure*}[t]
\centering
\includegraphics[width=0.9\textwidth]{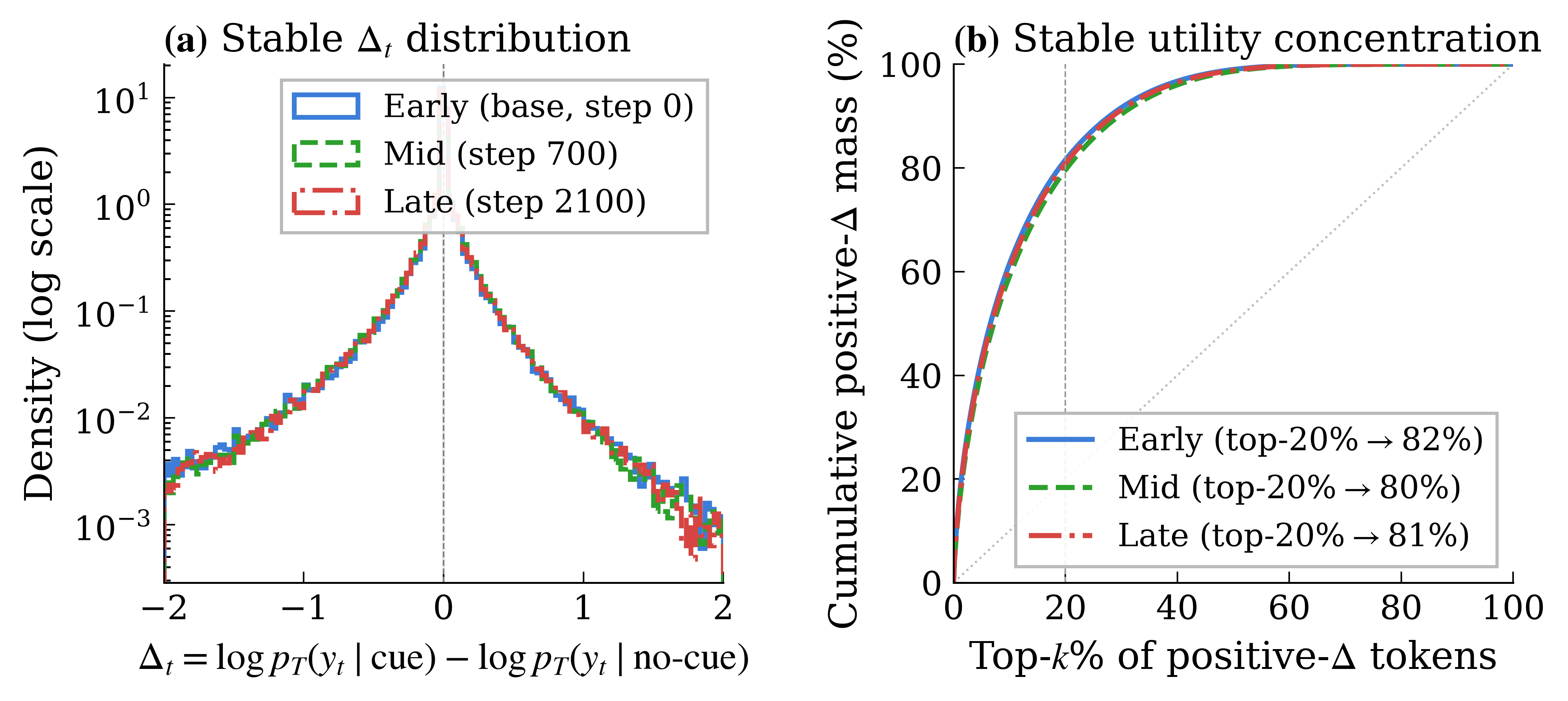}
\caption{Evolution of the cue-conditional log-likelihood ratio across student checkpoints. (a) The cue-versus-no-cue $\Delta_t$ distributions remain similar from the base model through steps 700 and 2100. (b) Positive-score concentration is also stable: the top $20\%$ of positive-$\Delta$ tokens carry approximately $80$--$82\%$ of the total positive mass at every stage.}
\label{fig:cue-delta-evolution}
\end{figure*}

To test whether the sparse cue-conditional pattern is limited to initialization, we repeat cue-versus-no-cue teacher scoring at three student checkpoints: the base model, step 700, and step 2100. Figure~\ref{fig:cue-delta-evolution} shows that the distributions remain closely aligned. The inactive-token fraction stays between $52.4\%$ and $56.5\%$, while the top $20\%$ of positive-$\Delta$ tokens consistently account for $80$--$82\%$ of the total positive mass. Thus, the sparse, heavy-tailed, and concentrated structure motivating ST-CueGate persists throughout optimization rather than appearing only at initialization.

\section{Negative Results: A Map of the Design Space}
For completeness we document teacher-condition and training variants that did not improve consistently over standard OPD or ST-CueGate. These results delineate the boundary of post-training and privilege design under our fixed inference protocol. All runs use the identical 25k backbone, optimizer, and evaluation protocol as the main paper unless noted.

\paragraph{Two-Axis Reward Extrapolation (AD-ExOPD).}
Decomposing the teacher's advantage over the base into a \emph{frame} axis (seeing more frames) and a \emph{capacity} axis (a larger model) with separate extrapolation gains $\lambda_t,\lambda_c$, we ran $\lambda_t{=}1.5,\lambda_c{=}1.0$. Validation oscillated between $50$--$65\%$ for the entire run (peak ${\sim}65\%$ at step 1600), never approaching the ${\sim}70\%$ that standard OPD reaches on the same data, and the extrapolation \emph{harmed} early convergence. Like single-axis extrapolation (ExOPD in the main text), pushing the student \emph{beyond} the teacher along an ungrounded direction does not help; adding a second such axis only destabilizes training.

\paragraph{Difficulty-Adaptive Distillation (DAD).}
As an ungrounded reweighting control, DAD assigns larger weights to responses on which student and teacher disagree. For response $y_i$, let $d_i=T_i^{-1}\sum_t|s_{i,t}-\tau_{i,t}|$. We first compute an unnormalized score and then normalize it within the batch:
\begin{align}
\widetilde{u}_i
&=\exp\!\left(\alpha_{\mathrm{DAD}}\,\mathrm{zscore}(d_i)\right), \notag\\
u_i
&=\mathrm{clip}\!\left(
\frac{\widetilde{u}_i}{\mathbb{E}_j[\widetilde{u}_j]},
\,0.2,\,3\right), \notag\\
A_{i,t}^{\mathrm{DAD}}
&=u_iA_{i,t}^{\mathrm{OPD}}.
\end{align}
With $\alpha_{\mathrm{DAD}}=1.0$, DAD reaches $84.27\%$ on StreamingBench but only $68.10\%$ on OVO-Bench (excl.\ HLD) and $62.93\%$ on Video-MME, providing no consistent improvement over standard OPD. This supports our decision to keep the main method focused on teacher-side grounded privilege rather than observation-agnostic difficulty weighting.

\paragraph{Mode-Dependent OPD Instability.}
Our recipe trains teacher and student both in \emph{thinking} mode. We tested the two additional pairings summarized by the training-mode figure in the main paper. \emph{Teacher-thinking / student-instruct}: with the student answering directly but the teacher scoring in thinking mode, validation falls from $69.3\%$ at step 0 to $26\%$ by step 100 and below $2\%$ by step 600. \emph{Both-instruct}: validation similarly drops from $69.9\%$ to $23.6\%$ by step 100. The failed runs produce very short responses and gradient-norm spikes of $70$--$120$, compared with $1$--$10$ in the stable both-thinking run. These observations are consistent with a short-trajectory scale hypothesis: sampled-token reverse-KL updates are supported by fewer token positions and may have higher effective variance. They do not isolate response length, however. Mode-specific initialization and policy distributions, teacher--student KL, prompt-induced distribution shift, response normalization, estimator variance, and importance-ratio clipping may also contribute. We therefore report the length-based account as a mechanistic hypothesis rather than an identified cause. We observe token-mean loss aggregation and no NaNs in these runs. Length-matched controls and multi-seed collapse rates are needed for causal identification.

\begin{figure*}[t]
\centering
\includegraphics[width=0.8\textwidth]{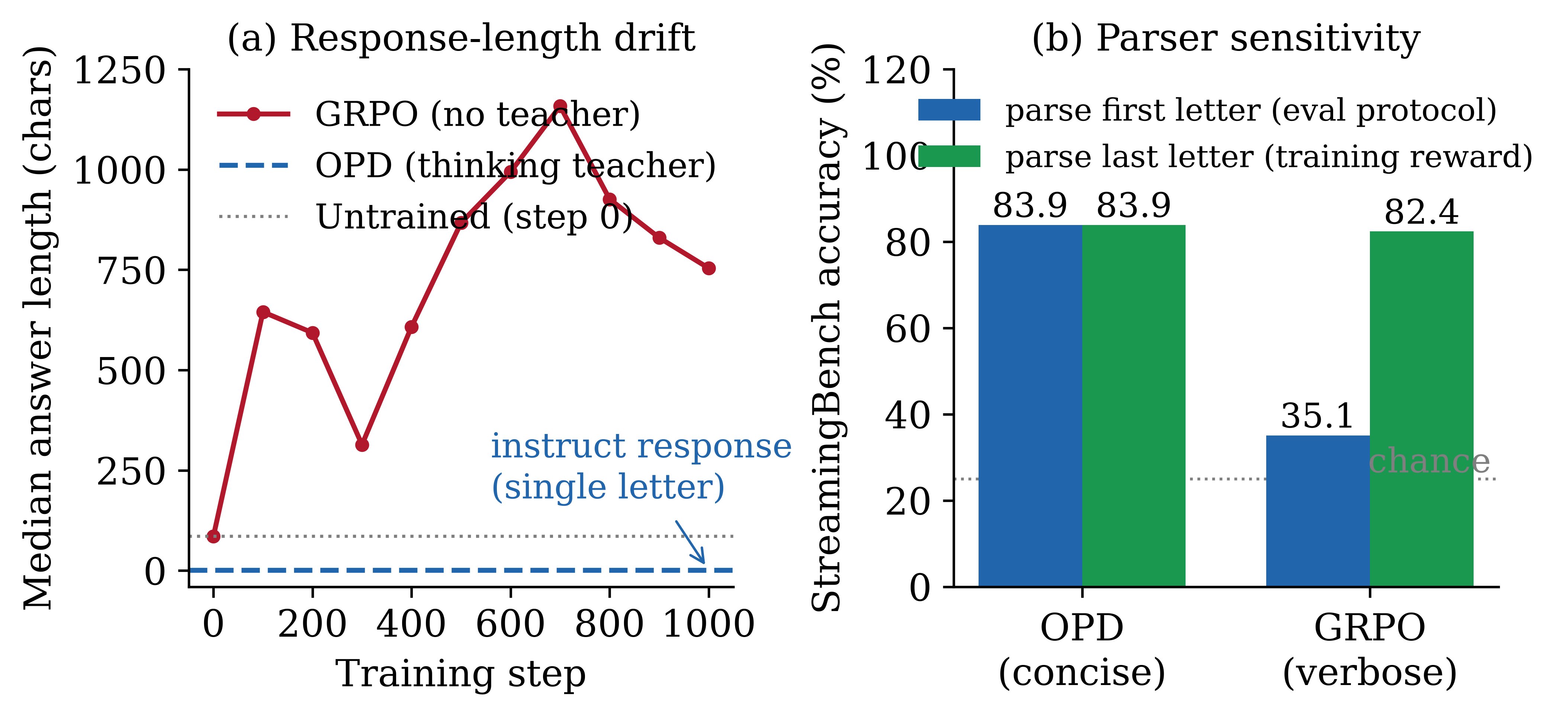}
\caption{Pure RL drifts from the deployed response format. (a) GRPO outputs grow from concise answers into long rationales, while OPD remains concise. (b) GRPO becomes sensitive to whether the evaluator reads the first or last answer, whereas the concise OPD model is parser-invariant. This gap diagnoses response-format drift under sparse reward.}
\label{fig:drift}
\end{figure*}

\paragraph{Teacher-Free RL in No-Think Mode (GRPO).}
Figure~\ref{fig:drift} plots the two format-drift diagnostics summarized in the main text: response-length growth over training, and sensitivity to whether the evaluator reads the first or the last answer. Consistent with that analysis, training GRPO directly in instruct mode diverges early, while GRPO in thinking mode plateaus near the untrained student. Because this comparison also changes mode-dependent generation and optimization behavior, we do not assign it the same short-trajectory mechanism without additional controls.

\paragraph{V-Zero Adaptation for Video.}
ST-CueGate uses the no-cue context as a nested reference: both teacher passes share the frames, question, and realized response, and differ only in cue presence. For the V-Zero adaptation~\citep{sun2026v}, we retain its paired-evidence gating principle and construct the negative evidence view by temporally shuffling the video. We also implemented a blacked-out negative view, but report the shuffled variant because it preserves appearance statistics while destroying only temporal order, which is the axis a streaming cue is meant to localize. This video variant scores below ST-CueGate on every reported axis (StreamingBench $83.5$--$84.0$, with OVO and Video-MME also lower), indicating that a shuffled-video reference is less effective than cue removal for our cue-conditioned streaming setting.

\paragraph{Asymmetric Frame Budget, Extended (AFD v2/v3).}
Beyond the AFD (t32/s8) configuration, we tried a denser teacher budget (fps~4, up to 64 frames; ``v2'') and a frame-superset design where the teacher's frames strictly contain the student's (``v3''). Both plateaued around ${\sim}60\%$ validation and never exceeded standard OPD on the same 25k data (which peaks at $70.6\%$); at matched training steps AFD trailed OPD by ${\sim}10$ points. Simply letting the teacher see more frames does not translate into better distilled supervision on our short-video data---the extra visual evidence is unavailable to the student at its own frame budget---which motivates measuring the cue-conditional teacher shift rather than supplying privilege blindly.

\section{Broader Impact and Deployment Considerations}
Streaming video models raise privacy considerations when deployed on live camera feeds. Our work uses public benchmark and instruction data and targets compute-efficient models, which lowers the energy cost of deployment. The HLD degradation we identify is safety-relevant even though HLD is distinct from streaming event recall: practitioners should be aware that distilled models can become more prone to answering unanswerable queries and should evaluate abstention explicitly before deployment.

\end{document}